\documentclass{article} %
\usepackage{iclr2024_conference,times}

\usepackage{amsmath,amsfonts,bm}

\def\eqref#1{equation~\ref{#1}}

\def\1{\bm{1}}

\def\rh{{\textnormal{h}}}
\def\ri{{\textnormal{i}}}

\def\rs{{\textnormal{s}}}

\def\rvh{{\mathbf{h}}}

\def\rvx{{\mathbf{x}}}

\def\rvz{{\mathbf{z}}}

\def\vmu{{\bm{\mu}}}
\def\vsigma{{\bm{\sigma}}}

\def\vtheta{{\bm{\theta}}}

\def\vh{{\bm{h}}}

\def\vx{{\bm{x}}}

\def\vz{{\bm{z}}}

\DeclareMathAlphabet{\mathsfit}{\encodingdefault}{\sfdefault}{m}{sl}
\SetMathAlphabet{\mathsfit}{bold}{\encodingdefault}{\sfdefault}{bx}{n}

\def\sX{{\mathbb{X}}}

\newcommand{\KL}{D_{\mathrm{KL}}}

\DeclareMathOperator*{\argmin}{arg\,min}

\newcommand{\mut}{\text{MI}} %
\newcommand{\lvl}{i}

\newtheorem{theorem}{Theorem}[section]

\newtheorem{lemma}[theorem]{Lemma}

\usepackage{hyperref}
\usepackage{commath}
\usepackage{url}
\usepackage{booktabs}
\usepackage{cleveref}
\usepackage{graphicx}
\usepackage{enumitem}
\usepackage{float}
\usepackage{xfrac}
\usepackage{algorithm}
\usepackage[noend]{algpseudocode}
\usepackage{etoolbox}
\usepackage{pdfpages}
\usepackage{subcaption}

\makeatletter
\newcommand*{\algrule}[1][\algorithmicindent]{%
  \makebox[#1][l]{%
    \hspace*{.2em}%
    \vrule height .75\baselineskip depth .25\baselineskip
  }
}

\newcount\ALG@printindent@tempcnta
\def\ALG@printindent{%
    \ifnum \theALG@nested>0%
    \ifx\ALG@text\ALG@x@notext%
    \else
    \unskip
    \ALG@printindent@tempcnta=1
    \loop
    \algrule[\csname ALG@ind@\the\ALG@printindent@tempcnta\endcsname]%
    \advance \ALG@printindent@tempcnta 1
    \ifnum \ALG@printindent@tempcnta<\numexpr\theALG@nested+1\relax
    \repeat
    \fi
    \fi
}
\patchcmd{\ALG@doentity}{\noindent\hskip\ALG@tlm}{\ALG@printindent}{}{\errmessage{failed to patch}}
\patchcmd{\ALG@doentity}{\item[]\nointerlineskip}{}{}{} %
\makeatother

\algrenewcommand\algorithmicrequire{\textbf{Input:}}
\algrenewcommand\algorithmicensure{\textbf{Output:}}
\algsetblock[Name]{Init}{6553}{0}{0cm}
\algdefaulttext[Name]{Into}
\newcommand{\Input}{\Require}

\newcommand{\rulesep}{\unskip\ \vrule\ }

\usepackage{amssymb}
\usepackage{mathtools}
\usepackage{wrapfig}
\usepackage{floatflt}
\usepackage{randtext}
\newlength{\twosubht}
\newsavebox{\twosubbox}

\title{How the level sampling process
impacts zero-shot generalisation in deep reinforcement learning}

\author{Samuel Garcin \\
University of Edinburgh\\
\texttt{\randomize{s.garcin@ed.ac.uk}}
\And
James Doran \\
University of Edinburgh\\
\texttt{\randomize{j.y.s.doran@sms.ed.ac.uk}}
\AND
Shangmin Guo \\
University of Edinburgh\\
\texttt{\randomize{s.guo@ed.ac.uk}}
\And
Christopher G. Lucas \\
University of Edinburgh\\
\texttt{\randomize{c.lucas@ed.ac.uk}}
\And
Stefano V. Albrecht \\
University of Edinburgh\\
\texttt{\randomize{s.albrecht@ed.ac.uk}}
}

\iclrfinalcopy %
\begin{document}

\maketitle
\vspace{-2em}

\begin{abstract}
A key limitation preventing the wider adoption of autonomous agents trained via deep reinforcement learning (RL) is their limited ability to generalise to new environments, even when these share similar characteristics with environments encountered during training. In this work, we investigate how a non-uniform sampling strategy of individual environment instances, or levels,  affects the zero-shot generalisation (ZSG) ability of RL agents, considering two failure modes: overfitting and over-generalisation. As a first step, we measure the mutual information ($\mut$) between the agent's internal representation and the set of training levels, which we find to be well-correlated to instance overfitting. In contrast to uniform sampling, adaptive sampling strategies prioritising levels based on their value loss are more effective at maintaining lower $\mut$, which provides a novel theoretical justification for this class of techniques. We then turn our attention to unsupervised environment design (UED) methods, which adaptively \textit{generate} new training levels and minimise $\mut$ more effectively than methods sampling from a fixed set. However, we find UED methods significantly \textit{shift} the training distribution, resulting in over-generalisation and worse ZSG performance over the distribution of interest. To prevent both instance overfitting and over-generalisation, we introduce \textit{self-supervised environment design} (SSED). SSED generates levels using a variational autoencoder, effectively reducing $\mut$ while minimising the shift with the distribution of interest, and leads to statistically significant improvements in ZSG over fixed-set level sampling strategies and UED methods.
\end{abstract}

\vspace{-.5em}
\section{introduction}\label{sec:intro}
\vspace{-.5em}
\begin{wrapfigure}[]{r}{0.5\textwidth}
    \centering
    \vspace{-.5cm}
    \begin{subfigure}{0.24\linewidth}
        \includegraphics[width=.95\linewidth]{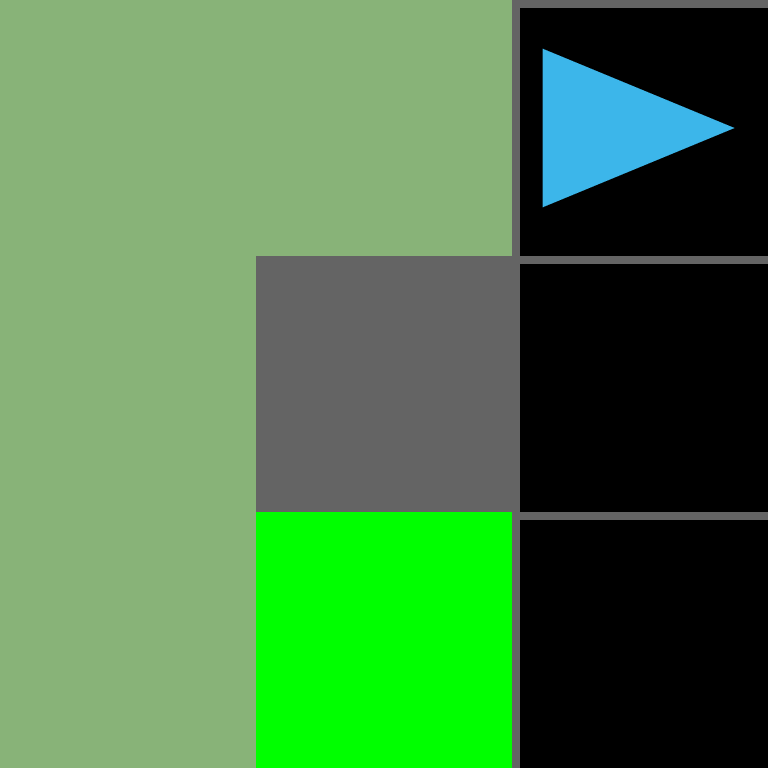}
        \caption{}
        \label{subfig:lvla}
    \end{subfigure}%
    \begin{subfigure}{0.24\linewidth}
        \includegraphics[width=.95\linewidth]{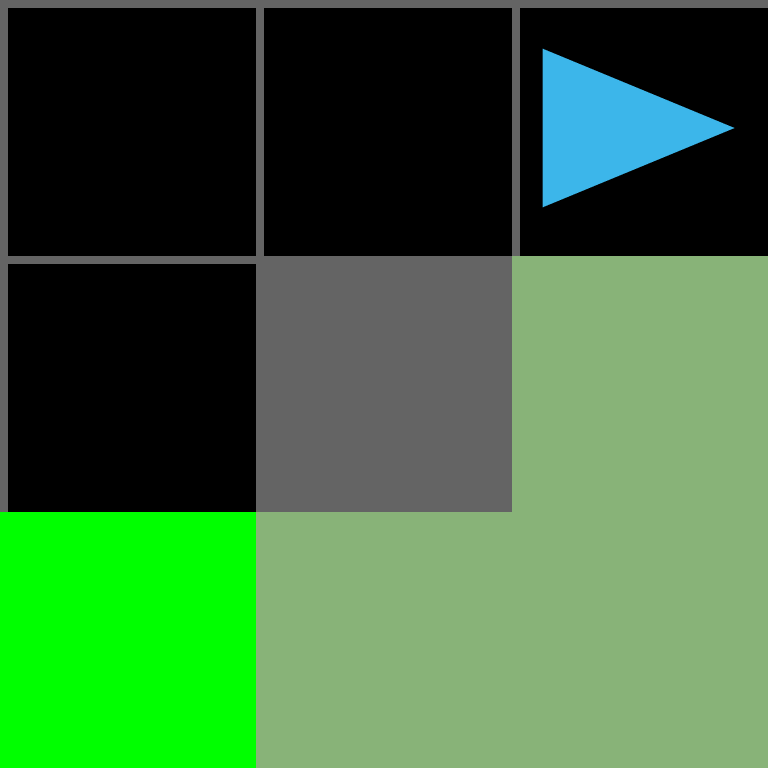}
        \caption{}
        \label{subfig:lvlb}
    \end{subfigure}%
        \begin{subfigure}{0.24\linewidth}
        \includegraphics[width=.95\linewidth]{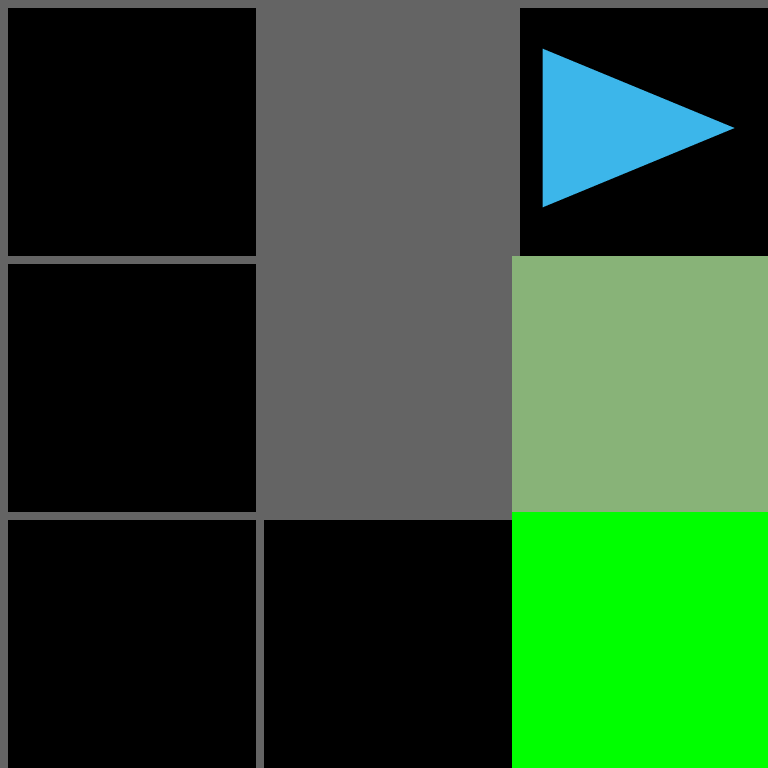}
        \caption{}
        \label{subfig:lvlc}
    \end{subfigure}%
    \hspace{-3pt}
    \rulesep
    \begin{subfigure}{0.24\linewidth}
        \includegraphics[width=0.95\linewidth]{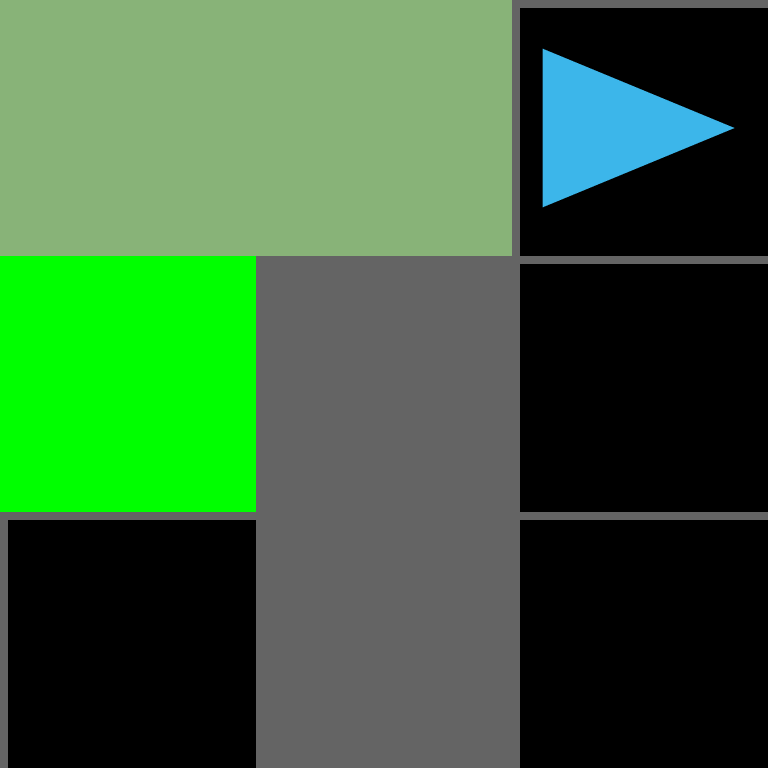}
        \caption{}
        \label{subfig:lvld}
    \end{subfigure}%
    \vspace{-.2cm}
    \caption{The agent (blue) must navigate to the goal (lime green) but cannot pass through walls (grey) and only observes tiles directly adjacent to itself. An agent trained over levels (a)-(c) will transfer zero-shot to level (d) if it has learnt a behavior adapted to the task semantics of following pale green tiles to reach the goal location.
    }
    \label{fig:CMDPex}
    \vspace{-.4cm}
\end{wrapfigure}
A central challenge facing modern reinforcement learning (RL) is learning policies capable of zero-shot transfer of learned behaviours to a wide range of environment settings. Prior applications of RL algorithms \citep{rubiks,CLrobloc,fastCLrobloc} indicate that strong zero-shot generalisation (ZSG) can be achieved through an adaptive sampling strategy over the set of environment instances available during training, which we refer to as the set of training \textit{levels}. However the relationship between ZSG and the level sampling process remains poorly understood. In this work, we draw novel connections between this process and the minimisation of an upper bound on the generalisation error derived by \cite{instance_invariant}, which depends on the \textit{mutual information} (MI) between the agent's internal representation and the identity of individual training levels.

An agent learning level-specific policies implies high MI between its internal representation and the level identities, and, in general, will not transfer zero-shot to new levels. To build an understanding of the relationship between MI and ZSG, consider the minimal gridworld navigation example in \Cref{fig:CMDPex}. A ``shortcut'' exists in level (a), and a model with high MI is able to first predict the level identity from its initial observation to learn an ensemble of level-specific policy optimal over the training set. When deployed on (d) the model will predict it is in (a) since under the agent's restricted field of view (a) and (d) share the same initial observation. As a result the agent will attempt to follow the (a)-specific policy, which will not transfer. We discover that the reduced generalisation error achieved by adaptive level sampling strategies over uniform sampling can be attributed to their effectiveness in reducing the MI between the agent's internal representation and the level identity. In particular, we find that strategies de-prioritising levels with low value loss, as proposed in prioritised level replay \citep[PLR,][]{PLR}, implicitly minimise mutual information as they avoid training on levels in which the value function is accurately estimated through level identification.

While some adaptive sampling strategies reduce the generalisation gap, their effectiveness is ultimately limited by the number of training levels. We propose \textit{Self-Supervised Environment Design} (SSED) which augments the set of training levels to further reduce generalisation error. We find training on an augmented set can negatively impact performance when the augmented set is not drawn from the same distribution as the training set. We show it induces a form of \textit{over-generalisation}, in which the agent learns to solve levels incompatible with the targeted task, and performs poorly at test time. There is therefore a trade-off between augmenting the training set to prevent instance-overfitting, i.e. to avoid learning level-specific policies, and ensuring that this augmented set comes from the same distribution to avoid distributional shift and over-generalisation. In our experiments, we show that SSED strikes this trade-off more effectively than other adaptive sampling and environment design methods. SSED achieves significant improvements in the agent's ZSG capabilities, reaching 1.25 times the returns of the next best baseline on held-out levels, and improving performance by two to three times on more difficult instantiations of the target task. %

\vspace{-.5em}
\section{related work}\label{sec:rw}
\vspace{-.5em}
\textbf{Buffer-free sampling strategies.} Domain randomisation \citep[DR,][]{DR, DR1997}, is one of the earliest proposed methods for improving the generalisation ability of RL agents by augmenting the set of available training levels, and does so by sampling uniformly between manually specified ranges of environment parameters. Subsequent contributions introduce an implicit prioritisation over the generated set by inducing a minimax return \citep[robust adversarial RL][]{Pinto2017RobustAdversarialRL} or a minimax regret \citep[unsupervised environment design (UED),][]{PAIRED} game between the agent and a \textit{level generator}, which are trained concurrently. These adversarial formulations prioritise levels in which the agent is currently performing poorly to encourage robust generalisation over the sampled set, with UED achieving better Nash equilibrium theoretical guarantees. CLUTR \citep{CLUTR} removes the need for domain-specific RL environments and improves sample efficiency by having the level generator operate within a low dimensional latent space of a generative model pre-trained on randomly sampled level parameters. However the performance and sample efficiency of these methods is poor when compared to a well calibrated DR implementation or to the buffer-based sampling strategies discussed next.

\textbf{Buffer-based sampling strategies.} Prioritised sampling is often applied to off-policy algorithms, where individual transitions in the replay buffer are prioritised \citep{Schaul2015PrioritizedER} or resampled with different goals in multi-goal RL \citep{HER,zhang2020automatic}. Prioritised Level Replay \citep[PLR,][]{PLR} instead affects the sampling process of \textit{future} experiences, and is thus applicable to both on- and off-policy algorithms. PLR maintains a buffer of training levels and empirically demonstrates that prioritising levels using a scoring function proportional to high value prediction loss results in better sample efficiency and improved ZSG performance. Robust PLR \citep[RPLR,][]{Robust-PLR} extends PLR to the UED setting, using DR as its level generation mechanism, whereas ACCEL \citep{ACCEL} gradually evolves new levels by performing random edits on high scoring levels in the buffer. SAMPLR \citep{SAMPLR} proposes to  eliminate the covariate shift induced by the prioritisation strategy by modifying \textit{individual transitions} using a second simulator that runs in parallel. However SAMPLR is only applicable to settings in which the level parameter distribution is provided, whereas SSED can approximate this distribution from a dataset of examples.

\textbf{Mutual-information minimisation in RL.} In prior work, mutual information has been minimised in order to mitigate instance-overfitting, either by learning an ensemble of policies \citep{instance_invariant,E-POMDP}, performing data augmentation on observations \citep{raileanu2021UCB-DrAC,Kostrikov2021ImageAllYouNeed}, an auxiliary objective \citep{mhairiCMID} or introducing information bottlenecks through selective noise injection on the agent model \citep{vib_actor_critic,coinrun}. In contrast, our work is the first to draw connections between mutual-information minimisation and adaptive level sampling and generation strategies.

\vspace{-.5em}
\section{preliminaries}\label{sec:bg}
\vspace{-.5em}
\textbf{Reinforcement learning.} We model an individual level as a Partially Observable Markov Decision Process (POMDP) $\langle A, O, S, \mathcal{T}, \Omega, R, p_0, \gamma \rangle$ where $A$ is the action space, $O$ is the observation space, $S$ is the set of states, $\mathcal{T} : S \times A \rightarrow \Delta (S)$ and $\Omega: S \rightarrow \Delta (O)$ are the transition and observation functions (we use $\Delta (\cdot)$ to indicate these functions map to distributions), $R:S\rightarrow \mathbb{R}$ is the reward function, $p_0(\rs)$ is the initial state distribution and $\gamma$ is the discount factor. We consider the episodic RL setting, in which the agent attempts to learn a policy $\pi$ maximising the expected discounted return $V^\pi (s_t) = \mathbb{E}_\pi[\sum^T_{\bar{t}=t}\gamma^{t - \bar{t}} r_t]$ over an episode terminating at timestep $T$, where $s_t$ and $r_t$ are the state and reward at step $t$. We use $V^\pi$ to refer to $V^\pi (s_0)$, the expected episodic returns taken from the first timestep of the episode. In this work, we focus on on-policy actor-critic algorithms \citep{A3C,DDPG,PPO} representing the agent policy $\pi_\vtheta$ and value estimate $\hat{V}_\vtheta$ with neural networks (in this paper we use $\vtheta$ to refer to model weights). The policy and value networks usually share an intermediate state representation $b_\vtheta(o_t)$ (or for recurrent architectures $b_\vtheta(H^o_t)$, $H^o_t = \{o_0,\cdots,o_t\}$ being the history of observations $o_i$).

\textbf{Contextual MDPs.} Following \cite{kirk2023survey}, we model the set of environment instances we aim to generalise over as a Contextual-MDP (CMDP) $\mathcal{M} = \langle A, O, S, \mathcal{T}, \Omega, R,  p_0(\rs|\rvx), \gamma, X_C, p(\rvx)\rangle$. The CMDP may be viewed as a POMDP, except that the reward, transition and observation functions now also depend on the \textit{context set} $X_C$ with associated distribution $p(\rvx)$, that is $\mathcal{T} : S \times X_C \times A \rightarrow \Delta (S)$, $\Omega: S \times X_C \rightarrow \Delta (O)$, $R:S \times X_C \rightarrow \mathbb{R}$. Each element $\vx \in X_C$ is not observable by the agent and instantiates a \textit{level} $\lvl_\vx$ of the CMDP with initial state distribution $p_0(\rs|\rvx)$. The optimal policy of the CMDP maximises $V^\pi_C = \mathbb{E}_{\rvx \sim p(\rvx)} [V^\pi_{\lvl_\vx}]$, with $V^\pi_{\lvl_\vx}$ referring to the expected return in level $\lvl_\vx$ instantiated by $\vx$ (we use $L$ to refer to a set of levels $i$ and $X$ to refer to a set of level parameters $\vx$). We assume access to a parametrisable simulator with parameter space $\sX$, with $X_C \subset \sX$. While prior work expects $X_C$ to correspond to all solvable levels in $\sX$, we consider the more general setting in which there may be more than one CMDP within $\sX$, whereas we aim to solve a specific target CMDP. We refer to levels with parameters $\vx \in X_C$ as \textit{in-context} and to levels with parameters $\vx \in \sX \setminus X_C$ as \textit{out-of-context}. As we show in our experiments, training on out-of-context levels can induce distributional shift and cause the agent to learn a different policy than the optimal CMDP policy.

\textbf{Generalisation bounds.} We start training with access to a limited set of level parameters $X_{\text{train}}\subset X_C$ sampled from $p(\rvx)$, and evaluate generalisation using a set of held-out level parameters $X_{\text{test}}$, also sampled from $p(\rvx)$. Using the unbiased value estimator lemma from \cite{instance_invariant},
\begin{lemma}\label{lem:unbiased_value_estimator}
    Given a policy $\pi$ and a set $L=\{\lvl_\vx | \vx \sim p(\rvx)\}_n$ of $n$ levels from a CMDP with context distribution $p(\rvx)$, we have $\forall H_t^o$  $(t < \infty)$ compatible with $L$ (that is the observation sequence $H_t^o$ occurs in $L$), $\mathbb{E}_{L|H_t^o}[V^\pi_{\lvl_\vx}(H_t^o)] = V^\pi_C(H_t^o)$, with $V^\pi_{\lvl_\vx}(H_t^o)$ being the expected returns under $\pi$ given observation history $H_t^o$ in a given level $\lvl_\vx$, and $V^\pi_C(H_t^o)$ being the expected returns across all possible occurrences of $H_t^o$ in the CMDP.
\end{lemma}
we can estimate the generalisation gap using a formulation reminiscent of supervised learning,
\begin{equation} \label{eq:GenGap}
    \text{GenGap}(\pi) \coloneqq \frac{1}{|X_{\text{train}}|} \sum_{\vx \in X_{\text{train}}} V^\pi_{\lvl_\vx} - \frac{1}{|X_{\text{test}}|}\sum_{\vx \in X_{\text{test}}} V^\pi_{\lvl_\vx}.
\end{equation}
Using this formulation, \cite{instance_invariant} extend generalisation results in the supervised setting \citep{MI_gengap_SL} to derive an upper bound for the $\text{GenGap}$. 
\begin{theorem}\label{th:gen_gap}
  For any CMDP such that $|V^\pi_C (H_t^o)| \leq D/2, \forall H_t^o, \pi$, then for any set of training levels $L$, and policy $\pi$
\begin{equation}
   \text{GenGap}(\pi) \leq \sqrt{\frac{2D^2}{|L|} \times \mut(L, \pi)}.
\end{equation}   
\end{theorem}
With $\mut (L, \pi) = \sum_{\lvl\in L} \mut(\lvl, \pi)$ being the mutual information between $\pi$ and the identity of each level $\lvl \in L$. In this work, we show that minimising the bound in \Cref{th:gen_gap} is an effective surrogate objective for reducing the $\text{GenGap}$. 

\textbf{Adaptive level sampling.} We study the connection between $\mut (L, \pi)$ and adaptive sampling strategies over $L$. PLR introduce a scoring function $\textbf{score}(\tau_i, \pi)$ compute level scores from a rollout trajectory $\tau_i$. Scores are used to define an adaptive sampling distribution over a level buffer $\Lambda$, with
\begin{equation}
    P_\Lambda = (1 - \rho) \cdot P_S + \rho \cdot P_R,
\end{equation}
where $P_S$ is a distribution parametrised by the level scores and $\rho$ is a coefficient mixing $P_S$ with a staleness distribution $P_R$ that promotes levels replayed less recently. \cite{PLR} experiment with different scoring functions, and empirically find that the scoring function based on the $\ell_1$-value loss $S^V_i = \textbf{score}(\tau_i, \pi) = (1 / |\tau_i|) \sum_{H_t^o\in\tau_i} |\hat{V} (H_t^o) - V^\pi_i (H_t^o) |$ incurs a significant reduction in the $\text{GenGap}$ at test time. 

In the remaining sections, we draw novel connections between the $\ell_1$-value loss prioritisation strategy and the minimisation of $\mut (L, \pi)$. We then introduce SSED, a level generation and sampling framework training the agent over an augmented set of levels. SSED jointly minimises $\mut (L, \pi)$ while increasing $|L|$ and as such is more effective at minimising the bound from \Cref{th:gen_gap}.
\vspace{-.5em}
\section{mutual-information minimisation under a fixed set of levels} \label{sec:procgen}
\vspace{-.5em}

We begin by considering the setting in which $L$ remains fixed. We make the following arguments: 1) as the contribution of each level to $\mut (L,\pi)$ is generally \textit{not uniform} across $L$ \textit{nor constant} over the course of training, an adaptive level sampling strategy yielding training data with low $\mut (L,\pi)$ can reduce the $\text{GenGap}$ over uniform sampling; 2) the value prediction objective promotes learning internal representations informative of the current level identity and causes overfitting; 3) de-prioritising levels with small value loss implicitly reshapes the training data distribution to yield smaller $\mut (L,\pi)$, reducing $\text{GenGap}$. We substantiate our arguments with a comparison of different sampling strategies in the Procgen benchmark \citep{procgen}.

\vspace{-.5em}
\subsection{maintaining low mutual information content via adaptive sampling}
\vspace{-.5em}
The following lemma enables us to derive an upper bound for $\mut (L,\pi)$ that can be approximated using the activations of the state representation shared between the actor and critic networks.
\begin{lemma} \label{th:mi_fixed_L}
    (proof in appendix) Given a set of training levels $L$ and an agent model $\pi = f \circ b$, where $b(H^o_t) = h_t$ is an intermediate state representation and $f$ is the policy head, we can bound $\mut (L,\pi \circ b)$ by $\mut (L,b)$, which in turn satisfies
    \begin{align}
        \mut (L,\pi) \leq \mut (L,b) &= \mathcal{H}(p(\ri)) + \sum_{i\in L} \int dh p(\rh, \ri) \log p(\ri|\rh) \\
        &\approx \mathcal{H}(p(\ri)) + \frac{1}{|B|} \sum_{(i, H_t^o) \in B} \log p(\ri|b(H_t^o)) \label{eq:empirical_approx2}
    \end{align}
    where $\mathcal{H}(p)$ is the entropy of $p$ and $B$ is a batch of trajectories collected from levels $i \in L$.
\end{lemma}
This result applies to any state representation function $b$, including the non-recurrent case where $b(H^o_t) = b(o_t), \forall (o, H^o_t) \in (O,O^{\otimes t})$. To remain consistent with the CMDP we must set $p(\ri)$ to $p(\rvx)$, making the entropy $\mathcal{H}(p(\ri))$ a constant. However the second term in \Cref{eq:empirical_approx2} %
depends on the representation $b$ learned under the training data. We hypothesise that minimising $\mut(L,b)$ in the training data is an effective data regularisation technique against instance-overfitting. We can isolate level-specific contributions to $\mut(L,b)$ as
\begin{equation}\label{eq:mi_2nd_term_decomposition}
    \sum_{(i, H_t^o) \in B} \log p(\ri|b(H_t^o)) = \sum_{i\in L} \sum_{H_t^o \in B_i} \log p(\ri|b(H_t^o)),
\end{equation}
where $B_i$ indicates the batch trajectories collected from level $i$. As sampled trajectories depend on the behavioral policy, and as the information being retained depends on $b$, each level's contribution to $\mut (L,b)$ is in general not constant over the course of training nor uniform across $L$. There should therefore exist adaptive distributions minimising $\mut (L,b)$ more effectively than uniform sampling.

\vspace{-.5em}
\subsection{on the effectiveness of value loss prioritisation strategies}
\vspace{-.5em}
From a representation learning perspective, the value prediction objective may be viewed as a self-supervised auxiliary objective shaping the intermediate state representation $b$. This additional signal is often necessary for learning, and motivates sharing $b$ across the policy and value networks. However, in a CMDP the value prediction loss 
\begin{equation} \label{eq:Lv}
    L_{V}(\vtheta) = \frac{1}{|B|} \sum_{(i, H_t^o) \in B} (\hat{V}_\vtheta(H_t^o) - V^\pi_i(H_t^o))^2
\end{equation}
uses level-specific functions $V^\pi_i$ as targets, which may be expressed as $V^\pi_i = V^\pi_C + v^\pi_i$, where $v^\pi_i$ is a component specific to $i$. While Lemma \ref{lem:unbiased_value_estimator} guarantees convergence to an unbiased estimator for $V^\pi_C$ when minimising \Cref{eq:Lv}, reaching zero training loss is only achievable by learning the level specific components $v^\pi_i$. Perfect value prediction requires learning an intermediate representation from which the current level $i$ is identifiable, which implies high $\mut (i, b)$. Conversely, we can characterise PLR's $\ell_1$-value loss sampling as a data regularisation technique minimising $\mut (L,b)$ when generating the training data. %
By de-prioritising levels with low $L_{V_i}$, PLR prevents the agent from generating training data for which its internal representation has started overfitting to.

\vspace{-.5em}
\subsection{comparing mutual information minimisation sampling strategies}
\vspace{-.5em}
We aim to establish how PLR under value loss scoring $S^V$ compares to a scoring function based on \Cref{eq:mi_2nd_term_decomposition}. We define this function as $S^{\mut}_i = \sum_{t=0}^T \log p_\vtheta(\ri|b(H_t^o))$, where $p_\vtheta$ is a linear classifier. We also introduce a secondary scoring strategy $S_2$ with associated distribution $P_{S_2}$, letting us \textit{mix} different sampling strategies and study their interaction. $P_\Lambda$ then becomes
\begin{equation} \label{eq:buffer_dist}
    P_\Lambda = (1 - \rho) \cdot ((1-\eta) \cdot  P_S + \eta \cdot P_{S_2}) + \rho \cdot P_R,
\end{equation}
with $\eta$ being a mixing parameter. We compare different sampling strategies in Procgen, a benchmark of 16 games designed to measure generalisation in RL. We train the PPO \citep{PPO} baseline employed in \citep{procgen}, which uses a non-recurrent intermediate representation $b_\vtheta (o_t)$ in the ``easy'' setting ($|L|=200$, $25M$ timesteps). We report a complete description of the experimental setup in \Cref{app:procgen}.

\begin{figure}[t]
    \centering
            \includegraphics[width=1\linewidth]{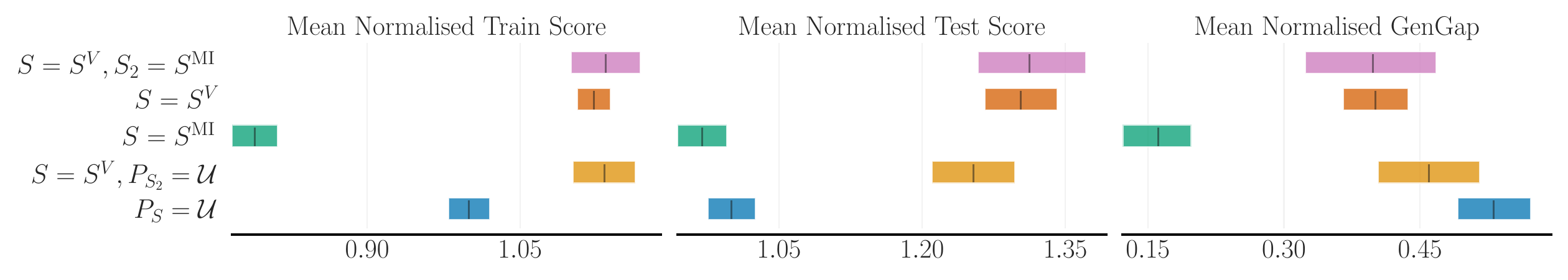}
    \caption{Mean aggregated train and test scores and $\text{GenGap}$ of different sampling strategies. Scores are normalised for each game using the mean score achieved by the uniform sampling strategy (we use the test set scores to normalise $\text{GenGap})$.
    }
    \label{fig:procgen_results}
\end{figure}

\Cref{fig:procgen_results}, compares value loss scoring ($S=S^V$), uniform sampling ($P_S=\mathcal{U}$), $\mathcal{U}(\cdot)$ being the uniform distribution, direct $\mut$ minimisation ($S=S^{\mut}$) as well as mixed strategies ($S=S^V, S_2=S^\mut$) and ($S=S^V, P_{S_2}=\mathcal{U}$). %
While ($S=S^\mut$) reduces $\text{GenGap}$ the most, the degradation it induces in the training performance outweigh its regularisation benefits. This result is consistent with \Cref{th:gen_gap} and Lemma \ref{th:mi_fixed_L}, as $\mut (L, b)$ bounds the GenGap and not the test returns.\footnote{Exclusively focusing on data regularisation can be problematic: in the most extreme case, destroying all information contained within the training data would guarantee $\mut(L,\pi) = \text{GenGap} = 0$ but it would also make the performance on the train and test sets equally bad.} On the other-hand, $(S=S^V)$ slightly improves training efficiency while reducing the $\text{GenGap}$. As denoted by its smaller $\text{GenGap}$ when compared to ($P_S=\mathcal{U}$), the improvements achieved by ($S=S^V$) are markedly stronger over the test set than for the train set, and indicate that the main driver behind the stronger generalisation performance is not a higher sample efficiency but a stronger regularisation. %
We tested different mixed strategies ($S=S^V, S_2=S^\mut$) using different $\eta$, and the best performing configuration (reported in \Cref{fig:procgen_results}) only achieves a marginal improvement over $(S=S^V)$ (on the other hand, mixing $S^V$ and uniform sampling ($S=S^V, P_{S_2}=\mathcal{U}$) noticeably reduces the test set performance). This implies that $(S=S^V)$ strikes a good balance between training efficiency and regularisation within the space of mutual information minimisation adaptive sampling strategies. In \Cref{app:procgen_results} we analyse the correlation between $\mut(L, b)$, the $\ell_1$-value loss and the $\text{GenGap}$ across all procgen games and methods tested. We find $\mut(L, b)$ to be positively correlated to the $\text{GenGap}$ $(p < 1\mathrm{e}{-34})$ and inversely correlated with the $\ell_1$-value loss $(p < 1\mathrm{e}{-16})$.

\vspace{-.5em}
\section{self-supervised environment design}\label{sec:ssed}
\vspace{-.5em}
We have established that certain adaptive sampling strategies effectively minimise $\mut (L,b)$, which in turn reduces $\text{GenGap}$. However our experiments in \Cref{sec:procgen,app:procgen_results,app:procgen_level_analysis} indicate $\text{GenGap}$ may still be significant when training the agent over a fixed level set, even with an adaptive sampling strategy. We now introduce SSED, a framework designed to more aggressively minimise the generalisation bound in \Cref{th:gen_gap} by jointly minimising $\mut (L, b)$ and increasing $|L|$. SSED does so by generating an augmented set of training levels $\tilde{L}\supset L$, while still employing an adaptive sampling strategy over the augmented set.

SSED shares UED's requirement of having access to a parametrisable simulator allowing the specification of levels through environment parameters $\vx$. In addition, we assume that we start with a limited set of level parameters $X_\text{train}$ sampled from $p(\rvx)$. SSED consists of two components: a \textit{generative phase}, in an augmented set $\tilde{X}$ is generated using a batch $X \sim \mathcal{U}(X_\text{train})$ and added to the buffer $\Lambda$, and a \textit{replay phase}, in which we use the adaptive distribution $P_\Lambda$ to sample levels from $\Lambda$. We alternate between the generative and replay phases, and only perform gradient updates on the agent during the replay phase. \Cref{alg:SSED} describes the full SSED pipeline, and we provide further details on each phase below.

\begin{algorithm}[]
\centering
\caption{Self-Supervised Environment Design}\label{alg:SSED}
\begin{algorithmic}[1]
\Input Pre-trained VAE encoder and decoder networks $\psi_{\vtheta_E}$, $\phi_{\vtheta_D}$, level parameters $X_\text{train}$, number of pairs $M$, number of interpolations per pair $K$
\State Initialise agent policy $\pi$ and level buffer $\Lambda$, adding level parameters in $X_\text{train}$ to $\Lambda$ 
\State Update $X_\text{train}$ with variational parameters $(\vmu_\rvz, \vsigma_\rvz)_n \gets \psi_{\vtheta_E}(\vx_n)$ \textbf{for} $\vx_n$ in $X_\text{train}$
\While{\textit{not converged}} 
    \State Sample batch $X$ using $P_\Lambda$ \Comment{Replay phase} %
    \For{$\vx$ in $X$}
        \State Collect rollouts $\tau$ from $i_\vx$ and compute scores $S, S_2$
        \State Update $\pi$ according to $\tau$
        \State Update scores $S, S_2$ of $\vx$ in $\Lambda$
    \EndFor
    \State Randomly sample $2M$ $(\vx, \vmu, \vsigma)$ from $X_\text{train}$ and arrange them into $M$ pairs. 
    \For {$((\vx, \vmu_\rvz, \vsigma_\rvz)_{i}, (\vx, \vmu_\rvz, \vsigma_\rvz)_{j})$ in pairs} \Comment{Generative phase}
        \State Compute $K$ interpolations $\{(\vmu_\rvz, \vsigma_\rvz)\}_{K}$ between $((\vx, \vmu_\rvz, \vsigma_\rvz)_{i}, (\vx, \vmu_\rvz, \vsigma_\rvz)_{j})$
        \For {$(\vmu_\rvz, \vsigma_\rvz)_k$ in $\{(\vmu_\rvz, \vsigma_\rvz)\}_{K}$}
            \State Sample embedding $\vz \sim \mathcal{N}(\vmu_\rvz, \vsigma_\rvz)$
            \State $\tilde{\vx} \gets \phi_{\vtheta_D}(\vz)$
            \State Collect $\pi$'s trajectory $\tau$ from $\vx$ and compute scores $S, S_2$
            \State Add $\langle \vx, S, S_2 \rangle$ to $\Lambda$ at $\argmin_{\{\Lambda \setminus X_\text{train}\}} S_2$ \textbf{if} $S_2 > \min_{\{\Lambda \setminus X_\text{train}\}} S_2$
        \EndFor
\EndFor
\EndWhile
\end{algorithmic}
\end{algorithm}

\vspace{-.5em}
\subsection{the generative phase}
\vspace{-.5em}
While SSED is not restricted to a particular approach to obtain $\tilde{X}$, we chose the VAE \citep{VAE, VAE-pmlr-v32-rezende14} due its ability to model the underlying training data distribution $p(\rvx)$ %
as stochastic realisations of a latent distribution $p(\rvz)$ via a generative model $p(\rvx \mid \rvz)$.
The model is pre-trained on $X_\text{train}$ by maximising the variational ELBO
\begin{equation} \label{eq:ELBO}
    \mathcal{L}_{\text{ELBO}} = \mathop{\mathbb{E}}_{\vx\sim p(\rvx)}\left\{\mathop{\mathbb{E}}_{\vz \sim q(\rvz|\rvx; \psi_{\vtheta_E})}[\log p(\rvx \mid \rvz; \phi_{\vtheta_D})] - \beta \KL(q(\rvz \mid \rvx;\psi_{\vtheta_E})\mid\mid p(\rvz))\right\},
\end{equation}
where $q(\rvz \mid \rvx; \psi_{\vtheta_E})$ is a variational approximation of an intractable model posterior distribution $p(\rvz \mid \rvx)$ and $\KL(\cdot \mid\mid \cdot)$ denotes the Kullback--Leibler divergence, which is balanced using the coefficient $\beta$, as proposed by \cite{betaVAE}. 
The generative $p(\rvx \mid \rvz; \phi_{\vtheta_D})$ and variational $q(\rvz \mid \rvx ; \psi_{\vtheta_E})$ models are parametrised via encoder and decoder networks $\psi_{\vtheta_E}$ and $\phi_{\vtheta_D}$.%

We use $p(\rvx; \phi_{\vtheta_D})$ to generate augmented level parameters $\tilde{\vx}$. As maximising \Cref{eq:ELBO} fits the VAE such that the marginal $p(\rvx; \phi_{\vtheta_D}) = \int p(\rvx \mid \rvz; \phi_{\vtheta_D}) p(\rvz) \dif \rvz$ approximates the data distribution $p(\rvx)$, and sampling from it limits distributional shift. This makes out-of-context levels less frequent, and we show in \Cref{sec:minigrid} that this aspect is key in enabling SSED-trained agents to outperform UED-agents. %
To improve the quality of the generated $\tilde{\vx}$, we interpolate in the latent space between the latent representations of pair of samples $(\vx_i, \vx_j) \sim X_\text{train}$ to obtain $\rvz$, instead of sampling from $p(\rvz)$, as proposed by \cite{white2016sampling}.
We evaluate the agent (without updating its weights) on the levels obtained from a batch of levels parameters $\tilde{X}$, adding to the buffer $\Lambda$ any level scoring higher than the lowest scoring generated level in $\Lambda$. We provide additional details on the architecture, hyperparameters and pre-training process in \Cref{app:vae_imp}.

\vspace{-.5em}
\subsection{the replay phase}
\vspace{-.5em}
All levels in $X_\text{train}$ originate from $p(\rvx)$ and are in-context, whereas generated levels, being obtained from an approximation of $p(\rvx)$, do not benefit from as strong of a guarantee. As training on out-of-context levels can significantly harm the agents' performance on the CMDP, we control the ratio between $X_\text{train}$ and augmented levels using \Cref{eq:buffer_dist} to define $P_\Lambda$. $P_S$ and $P_R$ only sample from $X_\text{train}$ levels, whereas $P_{S_2}$ supports the entire buffer. We set both $S_1$ and $S_2$ to score levels according to the $\ell_1$-value loss. We initialise the buffer $\Lambda$ to contain $X_\text{train}$ levels and gradually add generative phase levels over the course of training. A level gets added if $\Lambda$ is not full or if it scores higher than the lowest scoring level in the buffer. We only consider levels solved at least once during generative phase rollouts to ensure unsolvable levels do not get added in. We find out-of-context levels to be particularly harmful in the early stages of training, and reduce their frequency early on by linearly increasing the mixing parameter $\eta$ from 0 to 1 over the course of training.

\vspace{-.5em}
\section{experiments}\label{sec:minigrid}
\vspace{-.5em}
As it only permits level instantiation via providing a random seed, Procgen's level generation process is both uncontrollable and unobservable. Considering the seed space to be the level parameter space $\sX$ makes the level generation problem trivial as it is only possible to define a single CMDP with $X_C=\sX$ and $p(\rvx)=\mathcal{U(\sX)}$. Instead, we wish to demonstrate SSED's capability in settings where $X_C$ spans a (non-trivial) manifold in $\sX$, i.e. only specific parameter semantics will yield levels of the CMDP of interest. As such we pick Minigrid, a partially observable gridworld navigation domain \citep{gym_minigrid}. Minigrid levels can be instantiated via a parameter vector describing the locations, starting states and appearance of the objects in the grid. Despite its simplicity, Minigrid qualifies as a parametrisable simulator capable of instantiating multiple CMDPs. We define the context space of our target CMDP as spanning the layouts where the location of green ``moss'' tiles and orange ``lava'' tiles are respectively positively and negatively correlated to their distance to the goal location. 
We employ procedural generation to obtain a set $X_\text{train}$ of 512 level parameters, referring the reader to \Cref{fig:layouts_dataset} for a visualisation of levels from $X_\text{train}$, and to \Cref{app:cmdp_levelset} for extended details on the CMDP specification and level set generation process.

As the agent only observes its immediate surroundings and does not know the goal location a priori, the optimal CMDP policy is one that exploits the semantics shared by all levels in the CMDP, exploring first areas with high perceived moss density and avoiding areas with high lava density. Our CMDP coexists alongside a multitude of other potential CMDPs in the level space and some correspond to incompatible optimal policies (for example levels in which the correlation of moss and lava tiles with the goal is reversed). As such, it is important to maintain consistency with the CMDP semantics when generating new levels.

We compare SSED to multiple baselines, sorted in two sets. The first set of baselines is restricted to sample from $X_\text{train}$, and consists of uniform sampling ($\mathcal{U}$) and PLR with the $\ell_1$-value loss strategy. The second set incorporates a generative mechanism and as such are more similar to SSED. We consider domain randomisation (DR) \citep{DR} which generates levels by sampling uniformly between pre-determined ranges of parameters, RPLR \citep{Robust-PLR}, which combines PLR with DR used as its generator, and the current UED state-of-the-art, ACCEL \cite{ACCEL}, an extension of RPLR replacing DR by a generator making local edits to currently high scoring levels in the buffer. %
All experiments share the same PPO \citep{PPO} agent, which uses the LSTM-based architecture and hyperparameters reported in \cite{ACCEL}, training over 27k updates.
\vspace{-.5em}
\subsection{generalisation performance}
\begin{figure}[tbp]
    \begin{subfigure}[]{.08\linewidth}
    \centering
    \includegraphics[width=1\linewidth]{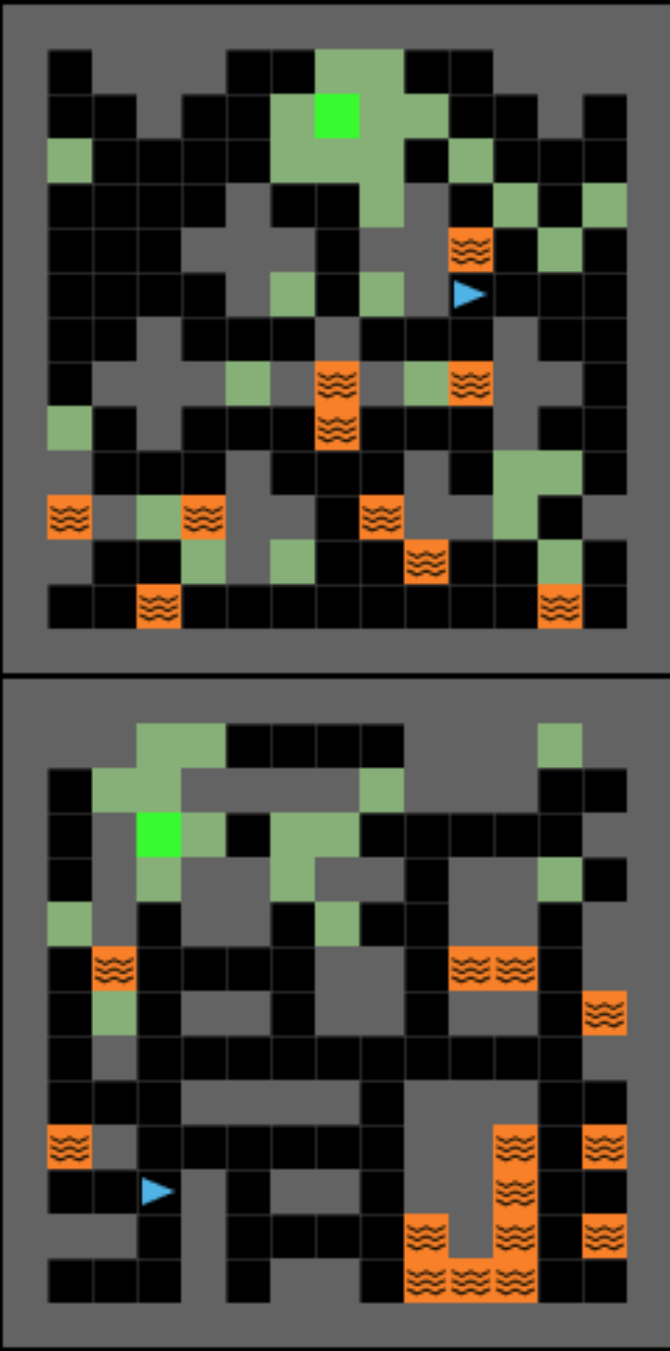}
  \end{subfigure}%
   \begin{subfigure}[]{.72\linewidth}
    \centering
    \includegraphics[width=1\linewidth]{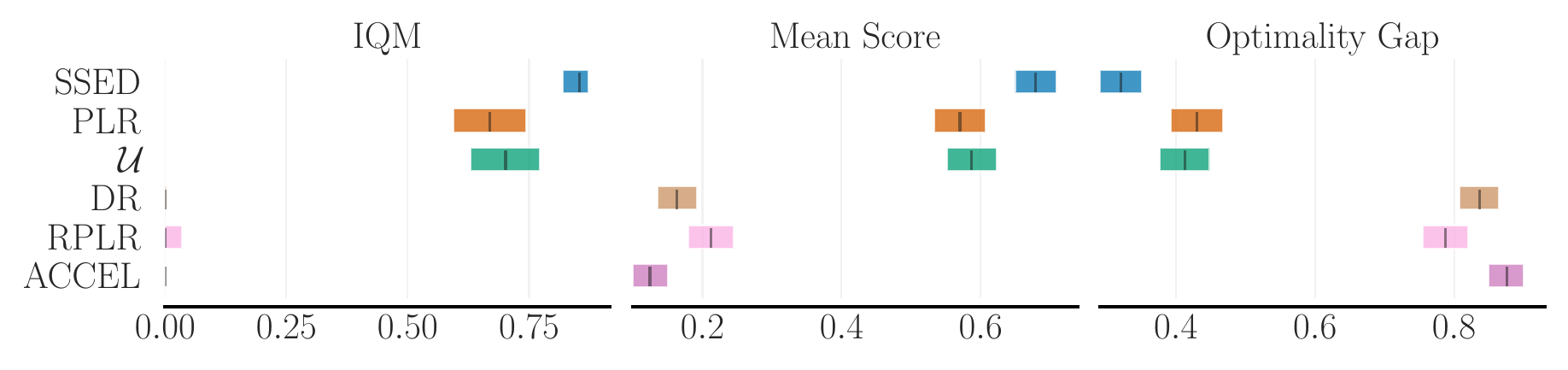}
  \end{subfigure}%
    \begin{subfigure}[]{.2\linewidth}
    \centering
    \includegraphics[width=1\linewidth]{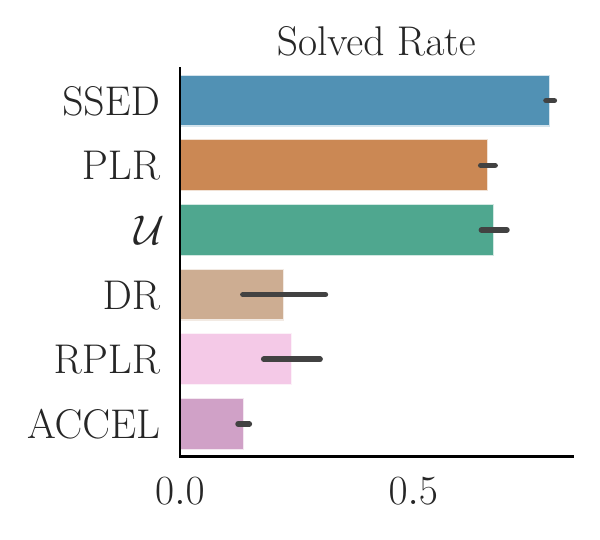}
    \end{subfigure}%
    
    \begin{subfigure}[]{.08\linewidth}
    \centering
    \includegraphics[width=1\linewidth]{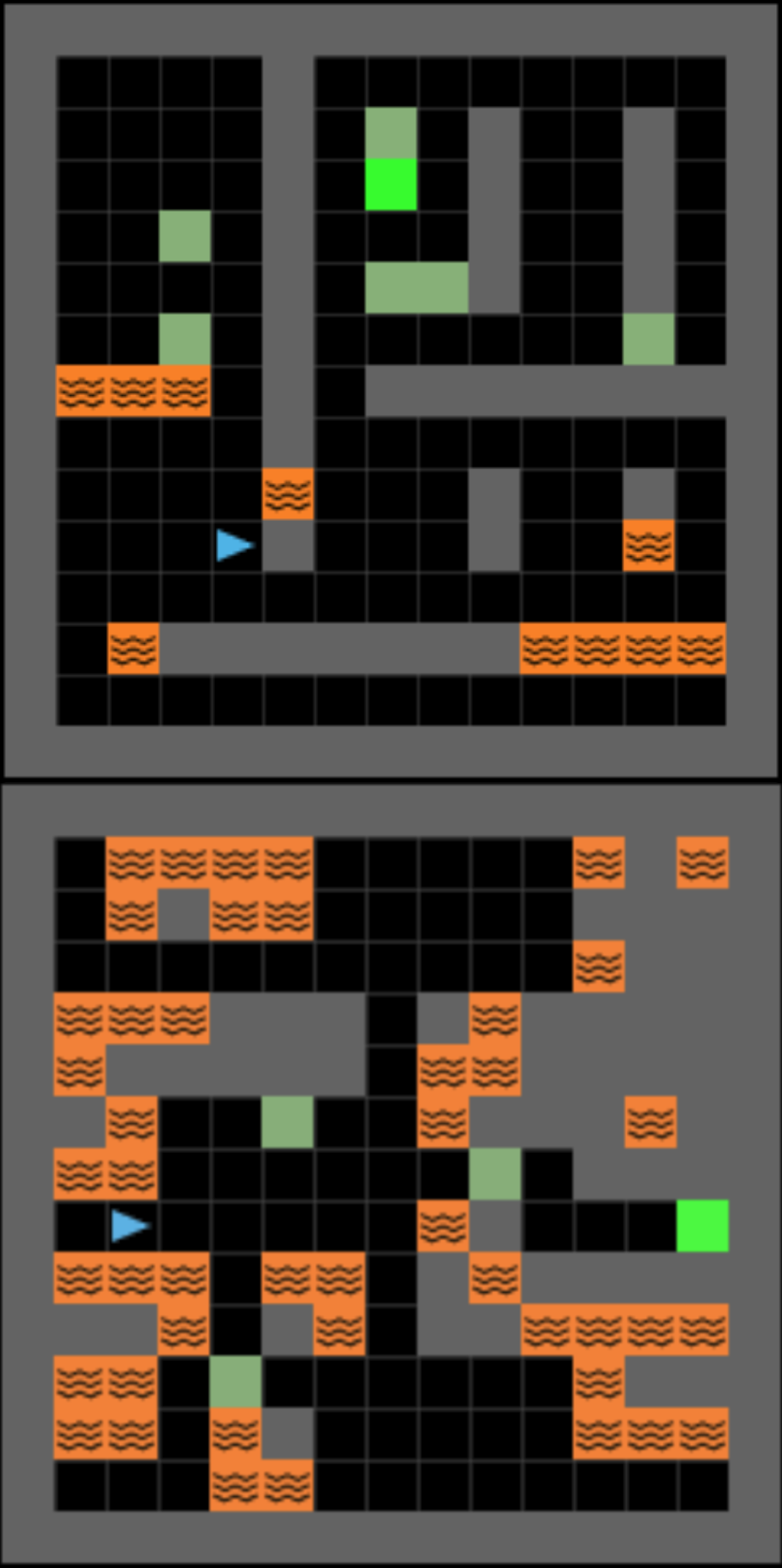}
  \end{subfigure}%
   \begin{subfigure}[]{.72\linewidth}
    \centering
    \includegraphics[width=1\linewidth]{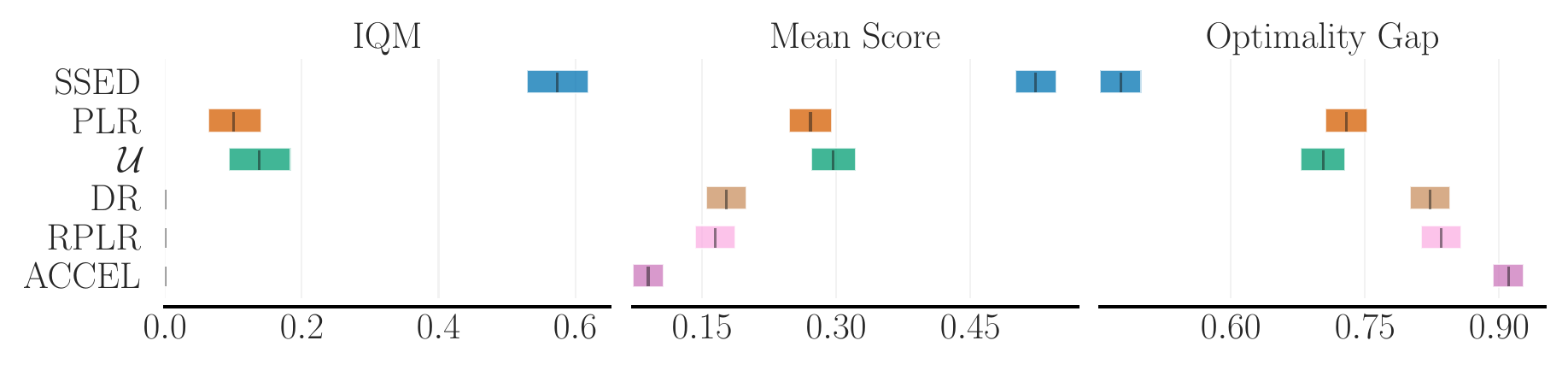}
  \end{subfigure}%
    \begin{subfigure}[]{.2\linewidth}
    \centering
    \includegraphics[width=1\linewidth]{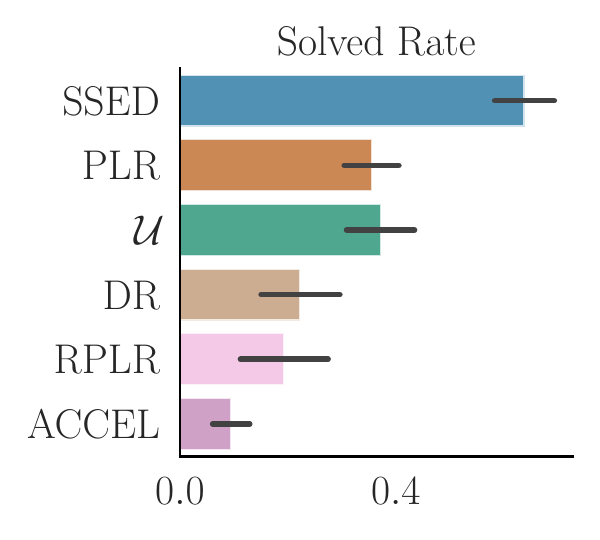}
    \end{subfigure}%
\caption{Center: aggregate test performance on 200 held-out levels from $X_\text{train}$ (top) and in-context edge cases (bottom). Right: zero-shot solved rate on the same levels, the bars indicate standard error for 3 training seeds (some example levels are provided for reference, refer to \Cref{app:cmdp_levelset} for additional details on our evaluation sets).}
\label{fig:zsg_test_set_and_edge_cases}
\vspace{-1.5em}
\end{figure}

\begin{figure}[tbp]
\centering
    \begin{subfigure}[]{.25\linewidth}
    \centering
    \includegraphics[width=1\linewidth]{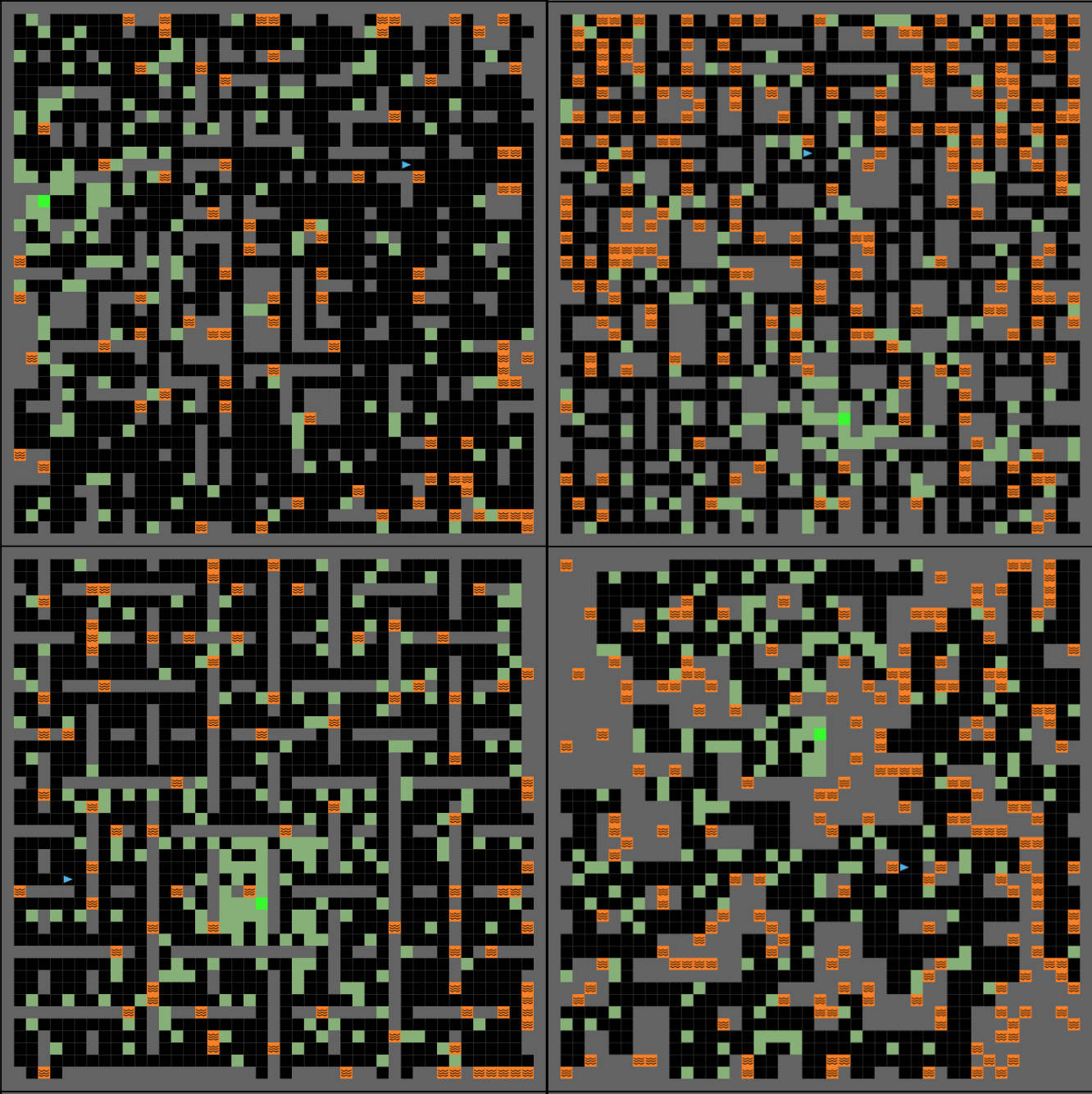}
  \end{subfigure}%
    \begin{subfigure}[]{.2\linewidth}
    \centering
    \includegraphics[width=1\linewidth]{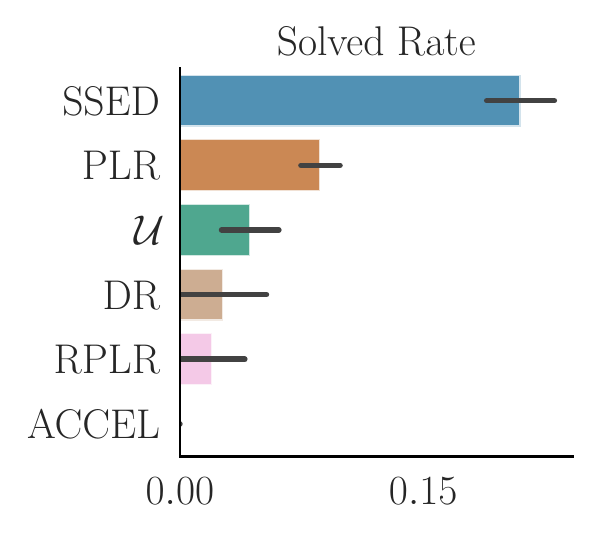}
    \end{subfigure}%
    \hspace{1cm}
    \begin{subfigure}[]{.3\linewidth}
    \centering
    \includegraphics[width=1\linewidth]{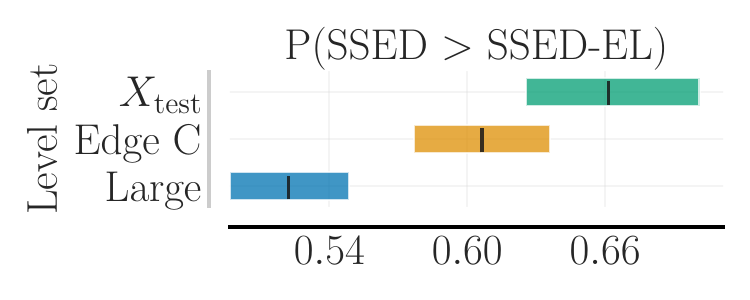}
    \end{subfigure}%
\caption{Left: zero-shot solved rate on a set of 100 levels with larger layouts. SSED's success rate is over twice PLR's, the next best performing method. Right: probability ($p<0.05$) of SSED achieving higher zero-shot returns than its ablation SSED-EL, evaluated in each level set. %
}
\label{fig:zsg_hard_and_ablation}
\vspace{-1.5em}
\end{figure}

\vspace{-.5em}
As shown in \Cref{fig:zsg_test_set_and_edge_cases}, SSED achieves statistically significant improvements in its IQM (inter-quantile mean), mean score, optimality gap and mean solved rate over other methods on held-out levels from the CMDP. SSED's ZSG performance on held-out levels from $X_\text{train}$ demonstrates it alleviates instance-overfitting while remaining consistent with the target CMDP. This is thanks to its generative model effectively approximating $p(\rvx)$, and to its mixed sampling strategy ensuring many training levels originate from $X_\text{train}$, which are guaranteed to be in-context. We next investigate whether SSED's level generation improves robustness to \textit{edge cases} which are in-context but would have a near zero likelihood of being in $X_\text{train}$ in a realistic setting. We model edge cases as levels matching the CMDP semantics but generated via different procedural generation parameters. We find SSED to be particularly dominant is this setting, achieving a solved rate and IQM respectively two- and four-times $\mathcal{U}$, the next best method, and a mean score 1.6 times PLR, the next best method for that metric. SSED is therefore capable of introducing additional diversity in the level set in a manner that remains semantically consistent with the CMDP. In \Cref{fig:zsg_hard_and_ablation}, we measure transfer to levels of increased complexity using a set of layouts 9 times larger in area than $X_\text{train}$ levels and which would be impossible to instantiate during training. We find that SSED performs over twice as well as the next best method in this setting. 

To better understand the importance of using a VAE as a generative model we introduce SSED-EL, a version of SSED replacing the VAE with ACCEL's level editing strategy. SSED-EL may be viewed as an SSED variant of ACCEL augmenting $X_\text{train}$ using a non-parametric generative method, or equivalently as an ablation of SSED that does not approximate $p(\rvx)$ and is therefore less grounded to the target CMDP. In \Cref{fig:zsg_hard_and_ablation}, we compare the two methods across level sets, and find that SSED improves more significantly over its ablation for level sets that are most similar to the original training set. This highlights the significance of being able to approximate $p(\rvx)$ through the VAE to avoid distributional shift, which we discuss next.

\vspace{-.5em}
\subsection{distributional shift and the over-generalisation gap}
\vspace{-.5em}
Despite poor test scores, the UED baselines achieve small $\text{GenGap}$ (as shown in \Cref{fig:gen_gaps}), as they perform poorly on both the test set and on $X_\text{train}$. Yet the fact they tend to perform well on the subset of $\sX$ spanning their own training distribution means that they have over-generalised to an out-of-context set. As such, we cannot qualify their poor performance on $X_C$ as a lack of capability but instead as a form of misgeneralisation not quantifiable by the $\text{GenGap}$, and which are reminiscent of goal misgeneralisation failure modes reported in \cite{goal_misgen_DRL, goal_misgen_deepmind}. Instead, we propose the \textit{over-generalisation gap} as a complementary metric, which we define as
\begin{equation} \label{eq:OverGenGap}
    \text{OverGap}(\pi) \coloneqq \sum_{\tilde{\vx} \in \Lambda} P_\Lambda(i_{\tilde{\vx}}) \cdot V^\pi_{\lvl_{\tilde{\vx}}} - \frac{1}{|X_{\text{train}}|} \sum_{\vx \in X_{\text{train}}} V^\pi_{\lvl_\vx}.
\end{equation}
Note that $\text{OverGap}$ compares the agent's performance with $X_\text{train}$ and as such is designed to measure over-generalisation induced by distributional shift.\footnote{ using $X_\text{test}$ would make $\text{OverGap}\equiv\text{GenGap}$ if $P_\Lambda = \mathcal{U}(X_\text{train})$, whereas it should be 0.} Based on further analysis conducted in \Cref{app:dist-shift}, high $\text{OverGap}$ coincides with the inclusion of out-of-context levels coupled with a significant shift in the level parameter distribution with respect to $p(\rvx)$, and we find that SSED is the only level generation method tested able to maintain both low distributional shift and $\text{OverGap}$.

\vspace{-.5em}
\section{conclusion}\label{sec:disc}
\vspace{-.5em}
In this work, we investigated the impact of the level sampling process on the ZSG capabilities of RL agents. We found adaptive sampling strategies are best understood as data regularisation techniques minimising the mutual information between the agent's internal representation and the identity of training levels. In doing so, these methods minimise an upper bound on the generalisation gap, and our experiments showed it to act as an effective proxy for reducing this gap in practice. This theoretical framing allowed us to understand the mechanisms behind the improved generalisation achieved by value loss prioritised level sampling, which had only been justified empirically in prior work. We then investigated the setting in which the set of training levels is not fixed and where the generalisation bound can be minimised by training over an augmented set. We proposed SSED, a level generation and sampling framework restricting level generation to an approximation of the underlying distribution of a starting set of level parameters. We showed that this restriction lets SSED mitigates the distributional shift induced by UED methods. By jointly minimising the generalisation and over-generalisation gaps, we demonstrated that SSED achieves strong generalisation performance on in-distribution test levels, while also being robust to in-context edge-cases.

In future work, we plan to investigate how SSED scales to more complex environments. In a practical setting, the level parameter space is often high dimensional, and levels are described by highly structured data corresponding to specific regions of the parameter space. Depending on the simulator used, level parameters may consist of sets of values, configuration files or any other modality specific to the simulator. For example, they could be 3D scans of indoor environments \citep{OpenRooms} or a vector map describing a city's road infrastructure \citep{argoverse2}, which are often costly to collect or prescribe manually, and thus are limited in supply. Augmenting the number of training environments is therefore likely to play a role in scaling up RL in a cost effective manner. Our experiments show that unsupervised environment generation is problematic even in gridworlds, whereas the SSED framework is designed to scale with the amount of data being provided.

Lastly, we are interested in further exploring the synergies between SSED and mutual information minimisation frameworks. SSED performs data augmentation uptream of level sampling, whereas \cite{PLR} report significant improvements in combining PLR with data augmentation on the agent's observations \citep{raileanu2021UCB-DrAC} and thus acting downstream of level sampling. There may be further synergies in combining mutual information minimisation techniques at different points of the level-to-agent information chain, which is an investigation we leave for future work.

\bibliography{iclr2024_conference}

\begin{thebibliography}{42}
\providecommand{\natexlab}[1]{#1}
\providecommand{\url}[1]{\texttt{#1}}
\expandafter\ifx\csname urlstyle\endcsname\relax
  \providecommand{\doi}[1]{doi: #1}\else
  \providecommand{\doi}{doi: \begingroup \urlstyle{rm}\Url}\fi

\bibitem[Agarwal et~al.(2021)Agarwal, Schwarzer, Castro, Courville, and
  Bellemare]{rliable}
Rishabh Agarwal, Max Schwarzer, Pablo~Samuel Castro, Aaron Courville, and
  Marc~G Bellemare.
\newblock Deep reinforcement learning at the edge of the statistical precipice.
\newblock \emph{Advances in Neural Information Processing Systems}, 2021.

\bibitem[Agostinelli et~al.(2019)Agostinelli, McAleer, Shmakov, and
  Baldi]{rubiks}
Forest Agostinelli, Stephen McAleer, Alexander Shmakov, and Pierre Baldi.
\newblock Solving the rubik's cube with deep reinforcement learning and search.
\newblock \emph{Nature Machine Intelligence}, 1\penalty0 (8):\penalty0
  356--363, 8 2019.
\newblock ISSN 2522-5839.
\newblock \doi{10.1038/s42256-019-0070-z}.
\newblock URL \url{https://doi.org/10.1038/s42256-019-0070-z}.

\bibitem[Andrychowicz et~al.(2017)Andrychowicz, Wolski, Ray, Schneider, Fong,
  Welinder, McGrew, Tobin, Pieter~Abbeel, and Zaremba]{HER}
Marcin Andrychowicz, Filip Wolski, Alex Ray, Jonas Schneider, Rachel Fong,
  Peter Welinder, Bob McGrew, Josh Tobin, OpenAI Pieter~Abbeel, and Wojciech
  Zaremba.
\newblock Hindsight experience replay.
\newblock \emph{Advances in neural information processing systems}, 30, 2017.

\bibitem[Azad et~al.(2023)Azad, Gur, Emhoff, Alexis, Faust, Abbeel, and
  Stoica]{CLUTR}
Abdus~Salam Azad, Izzeddin Gur, Jasper Emhoff, Nathaniel Alexis, Aleksandra
  Faust, Pieter Abbeel, and Ion Stoica.
\newblock Clutr: Curriculum learning via unsupervised task representation
  learning.
\newblock In \emph{International Conference on Machine Learning}, pp.\
  1361--1395. PMLR, 2023.

\bibitem[Bellemare et~al.(2015)Bellemare, Naddaf, Veness, and Bowling]{ALE}
Marc~G. Bellemare, Yavar Naddaf, Joel Veness, and Michael Bowling.
\newblock The arcade learning environment: An evaluation platform for general
  agents (extended abstract).
\newblock In \emph{IJCAI}, 2015.

\bibitem[Bertran et~al.(2020)Bertran, Martinez, Phielipp, and
  Sapiro]{instance_invariant}
Martin Bertran, Natalia Martinez, Mariano Phielipp, and Guillermo Sapiro.
\newblock Instance based generalization in reinforcement learning.
\newblock \emph{CoRR}, abs/2011.01089, 2020.
\newblock URL \url{https://arxiv.org/abs/2011.01089}.

\bibitem[Chevalier-Boisvert et~al.(2018)Chevalier-Boisvert, Willems, and
  Pal]{gym_minigrid}
Maxime Chevalier-Boisvert, Lucas Willems, and Suman Pal.
\newblock Minimalistic gridworld environment for openai gym.
\newblock \url{https://github.com/maximecb/gym-minigrid}, 2018.

\bibitem[Cobbe et~al.(2019)Cobbe, Klimov, Hesse, Kim, and Schulman]{coinrun}
Karl Cobbe, Oleg Klimov, Chris Hesse, Taehoon Kim, and John Schulman.
\newblock Quantifying generalization in reinforcement learning.
\newblock In Kamalika Chaudhuri and Ruslan Salakhutdinov (eds.),
  \emph{Proceedings of the 36th International Conference on Machine Learning},
  volume~97 of \emph{Proceedings of Machine Learning Research}, pp.\
  1282--1289. PMLR, 09--15 Jun 2019.
\newblock URL \url{https://proceedings.mlr.press/v97/cobbe19a.html}.

\bibitem[Cobbe et~al.(2020)Cobbe, Hesse, Hilton, and Schulman]{procgen}
Karl Cobbe, Chris Hesse, Jacob Hilton, and John Schulman.
\newblock Leveraging procedural generation to benchmark reinforcement learning.
\newblock In \emph{International conference on machine learning}, pp.\
  2048--2056. PMLR, 2020.

\bibitem[Dennis et~al.(2020)Dennis, Jaques, Vinitsky, Bayen, Russell, Critch,
  and Levine]{PAIRED}
Michael Dennis, Natasha Jaques, Eugene Vinitsky, Alexandre Bayen, Stuart
  Russell, Andrew Critch, and Sergey Levine.
\newblock Emergent complexity and zero-shot transfer via unsupervised
  environment design.
\newblock \emph{NIPS}, 2020.

\bibitem[Di~Langosco et~al.(2022)Di~Langosco, Koch, Sharkey, Pfau, and
  Krueger]{goal_misgen_DRL}
Lauro~Langosco Di~Langosco, Jack Koch, Lee~D Sharkey, Jacob Pfau, and David
  Krueger.
\newblock Goal misgeneralization in deep reinforcement learning.
\newblock In \emph{International Conference on Machine Learning}, pp.\
  12004--12019. PMLR, 2022.

\bibitem[Dunion et~al.(2023)Dunion, McInroe, Luck, Hanna, and
  Albrecht]{mhairiCMID}
Mhairi Dunion, Trevor McInroe, Kevin~Sebastian Luck, Josiah~P. Hanna, and
  Stefano~V. Albrecht.
\newblock Conditional mutual information for disentangled representations in
  reinforcement learning, 2023.

\bibitem[Ghosh et~al.(2021)Ghosh, Rahme, Kumar, Zhang, Adams, and
  Levine]{E-POMDP}
Dibya Ghosh, Jad Rahme, Aviral Kumar, Amy Zhang, Ryan~P. Adams, and Sergey
  Levine.
\newblock Why generalization in {RL} is difficult: Epistemic pomdps and
  implicit partial observability.
\newblock \emph{CoRR}, abs/2107.06277, 2021.
\newblock URL \url{https://arxiv.org/abs/2107.06277}.

\bibitem[Gumin(2016)]{WFC}
M.~Gumin.
\newblock Wave function collapse algorithm.
\newblock \url{https://github.com/mxgmn/}, 2016.

\bibitem[Higgins et~al.(2017)Higgins, Matthey, Pal, Burgess, Glorot, Botvinick,
  Mohamed, and Lerchner]{betaVAE}
Irina Higgins, Lo{\"i}c Matthey, Arka Pal, Christopher~P. Burgess, Xavier
  Glorot, Matthew~M. Botvinick, Shakir Mohamed, and Alexander Lerchner.
\newblock beta-vae: Learning basic visual concepts with a constrained
  variational framework.
\newblock In \emph{ICLR}, 2017.

\bibitem[Igl et~al.(2019)Igl, Ciosek, Li, Tschiatschek, Zhang, Devlin, and
  Hofmann]{vib_actor_critic}
Maximilian Igl, Kamil Ciosek, Yingzhen Li, Sebastian Tschiatschek, Cheng Zhang,
  Sam Devlin, and Katja Hofmann.
\newblock Generalization in reinforcement learning with selective noise
  injection and information bottleneck.
\newblock In \emph{Neural Information Processing Systems}, 2019.
\newblock URL \url{https://api.semanticscholar.org/CorpusID:202778414}.

\bibitem[Jakobi(1997)]{DR1997}
Nick Jakobi.
\newblock Evolutionary robotics and the radical envelope-of-noise hypothesis.
\newblock \emph{Adaptive Behavior}, 6:\penalty0 325 -- 368, 1997.

\bibitem[Jiang et~al.(2021{\natexlab{a}})Jiang, Dennis, Parker-Holder,
  Foerster, Grefenstette, and Rockt\"{a}schel]{Robust-PLR}
Minqi Jiang, Michael Dennis, Jack Parker-Holder, Jakob Foerster, Edward
  Grefenstette, and Tim Rockt\"{a}schel.
\newblock Replay-guided adversarial environment design.
\newblock In M.~Ranzato, A.~Beygelzimer, Y.~Dauphin, P.S. Liang, and J.~Wortman
  Vaughan (eds.), \emph{Advances in Neural Information Processing Systems},
  volume~34, pp.\  1884--1897. Curran Associates, Inc., 2021{\natexlab{a}}.
\newblock URL
  \url{https://proceedings.neurips.cc/paper/2021/file/0e915db6326b6fb6a3c56546980a8c93-Paper.pdf}.

\bibitem[Jiang et~al.(2021{\natexlab{b}})Jiang, Grefenstette, and
  Rockt{\"a}schel]{PLR}
Minqi Jiang, Edward Grefenstette, and Tim Rockt{\"a}schel.
\newblock Prioritized level replay.
\newblock \emph{ArXiv}, abs/2010.03934, 2021{\natexlab{b}}.

\bibitem[Jiang et~al.(2022)Jiang, Dennis, Parker-Holder, Lupu, K{\"u}ttler,
  Grefenstette, Rockt{\"a}schel, and Foerster]{SAMPLR}
Minqi Jiang, Michael Dennis, Jack Parker-Holder, Andrei Lupu, Heinrich
  K{\"u}ttler, Edward Grefenstette, Tim Rockt{\"a}schel, and Jakob Foerster.
\newblock Grounding aleatoric uncertainty in unsupervised environment design.
\newblock \emph{arXiv preprint arXiv:2207.05219}, 2022.

\bibitem[Kingma \& Welling(2014)Kingma and Welling]{VAE}
Diederik~P Kingma and Max Welling.
\newblock Auto-encoding variational bayes.
\newblock \emph{ICLR}, 2014.

\bibitem[Kirk et~al.(2023)Kirk, Zhang, Grefenstette, and
  Rockt{\"a}schel]{kirk2023survey}
Robert Kirk, Amy Zhang, Edward Grefenstette, and Tim Rockt{\"a}schel.
\newblock A survey of zero-shot generalisation in deep reinforcement learning.
\newblock \emph{Journal of Artificial Intelligence Research}, 76:\penalty0
  201--264, 2023.

\bibitem[Kostrikov et~al.(2021)Kostrikov, Yarats, and
  Fergus]{Kostrikov2021ImageAllYouNeed}
Ilya Kostrikov, Denis Yarats, and Rob Fergus.
\newblock Image augmentation is all you need: Regularizing deep reinforcement
  learning from pixels.
\newblock \emph{ArXiv}, abs/2004.13649, 2021.

\bibitem[Krizhevsky et~al.(2012)Krizhevsky, Sutskever, and Hinton]{CNNimagenet}
Alex Krizhevsky, Ilya Sutskever, and Geoffrey~E Hinton.
\newblock Imagenet classification with deep convolutional neural networks.
\newblock \emph{Advances in neural information processing systems}, 25, 2012.

\bibitem[Lee et~al.(2020)Lee, Hwango, et~al.]{CLrobloc}
J.~Lee, J.~Hwango, et~al.
\newblock Learning quadrupedal locomotion over challenging terrain.
\newblock \emph{Science Robotics}, 2020.

\bibitem[Li et~al.(2021)Li, Yu, Sang, Wang, Song, Liu, Yeh, Zhu, Gundavarapu,
  Shi, Bi, Yu, Xu, Sunkavalli, Hasan, Ramamoorthi, and Chandraker]{OpenRooms}
Z.~Li, T.~Yu, S.~Sang, S.~Wang, M.~Song, Y.~Liu, Y.~Yeh, R.~Zhu,
  N.~Gundavarapu, J.~Shi, S.~Bi, H.~Yu, Z.~Xu, K.~Sunkavalli, M.~Hasan,
  R.~Ramamoorthi, and M.~Chandraker.
\newblock Openrooms: An open framework for photorealistic indoor scene
  datasets.
\newblock In \emph{2021 IEEE/CVF Conference on Computer Vision and Pattern
  Recognition (CVPR)}, pp.\  7186--7195, Los Alamitos, CA, USA, jun 2021. IEEE
  Computer Society.
\newblock \doi{10.1109/CVPR46437.2021.00711}.
\newblock URL
  \url{https://doi.ieeecomputersociety.org/10.1109/CVPR46437.2021.00711}.

\bibitem[Lillicrap et~al.(2016)Lillicrap, Hunt, Pritzel, Heess, Erez, Tassa,
  Silver, and Wierstra]{DDPG}
Timothy~P. Lillicrap, Jonathan~J. Hunt, Alexander Pritzel, Nicolas Manfred~Otto
  Heess, Tom Erez, Yuval Tassa, David Silver, and Daan Wierstra.
\newblock Continuous control with deep reinforcement learning.
\newblock \emph{CoRR}, abs/1509.02971, 2016.

\bibitem[Mnih et~al.(2016)Mnih, Badia, Mirza, Graves, Lillicrap, Harley,
  Silver, and Kavukcuoglu]{A3C}
Volodymyr Mnih, Adri{\`a}~Puigdom{\`e}nech Badia, Mehdi Mirza, Alex Graves,
  Timothy~P. Lillicrap, Tim Harley, David Silver, and Koray Kavukcuoglu.
\newblock Asynchronous methods for deep reinforcement learning.
\newblock In \emph{ICML}, 2016.

\bibitem[Parker-Holder et~al.(2022)Parker-Holder, Jiang, Dennis, Samvelyan,
  Foerster, Grefenstette, and Rocktaschel]{ACCEL}
Jack Parker-Holder, Minqi Jiang, Michael Dennis, Mikayel Samvelyan, Jakob~N.
  Foerster, Edward Grefenstette, and Tim Rocktaschel.
\newblock Evolving curricula with regret-based environment design.
\newblock \emph{ArXiv}, abs/2203.01302, 2022.

\bibitem[Pinto et~al.(2017)Pinto, Davidson, Sukthankar, and
  Gupta]{Pinto2017RobustAdversarialRL}
Lerrel Pinto, James Davidson, Rahul Sukthankar, and Abhinav~Kumar Gupta.
\newblock Robust adversarial reinforcement learning.
\newblock In \emph{ICML}, 2017.

\bibitem[Raileanu et~al.(2021)Raileanu, Goldstein, Yarats, Kostrikov, and
  Fergus]{raileanu2021UCB-DrAC}
Roberta Raileanu, Maxwell Goldstein, Denis Yarats, Ilya Kostrikov, and Rob
  Fergus.
\newblock Automatic data augmentation for generalization in reinforcement
  learning.
\newblock \emph{Advances in Neural Information Processing Systems},
  34:\penalty0 5402--5415, 2021.

\bibitem[Rezende et~al.(2014)Rezende, Mohamed, and
  Wierstra]{VAE-pmlr-v32-rezende14}
Danilo~Jimenez Rezende, Shakir Mohamed, and Daan Wierstra.
\newblock Stochastic backpropagation and approximate inference in deep
  generative models.
\newblock In Eric~P. Xing and Tony Jebara (eds.), \emph{Proceedings of the 31st
  International Conference on Machine Learning}, volume~32 of \emph{Proceedings
  of Machine Learning Research}, pp.\  1278--1286, Bejing, China, 22--24 Jun
  2014. PMLR.
\newblock URL \url{https://proceedings.mlr.press/v32/rezende14.html}.

\bibitem[Rudin et~al.(2021)Rudin, Hoeller, Reist, and Hutter]{fastCLrobloc}
Nikita Rudin, David Hoeller, Philipp Reist, and Marco Hutter.
\newblock Learning to walk in minutes using massively parallel deep
  reinforcement learning.
\newblock \emph{arXiv preprint arXiv:2109.11978}, 2021.

\bibitem[Schaul et~al.(2015)Schaul, Quan, Antonoglou, and
  Silver]{Schaul2015PrioritizedER}
Tom Schaul, John Quan, Ioannis Antonoglou, and David Silver.
\newblock Prioritized experience replay.
\newblock \emph{CoRR}, abs/1511.05952, 2015.
\newblock URL \url{https://api.semanticscholar.org/CorpusID:13022595}.

\bibitem[Schulman et~al.(2017)Schulman, Wolski, Dhariwal, Radford, and
  Klimov]{PPO}
John Schulman, Filip Wolski, Prafulla Dhariwal, Alec Radford, and Oleg Klimov.
\newblock Proximal policy optimization algorithms.
\newblock \emph{arXiv}, 2017.

\bibitem[Shah et~al.(2022)Shah, Varma, Kumar, Phuong, Krakovna, Uesato, and
  Kenton]{goal_misgen_deepmind}
Rohin Shah, Vikrant Varma, Ramana Kumar, Mary Phuong, Victoria Krakovna,
  Jonathan Uesato, and Zac Kenton.
\newblock Goal misgeneralization: Why correct specifications aren't enough for
  correct goals.
\newblock \emph{arXiv preprint arXiv:2210.01790}, 2022.

\bibitem[Tobin et~al.(2017)Tobin, Fong, Ray, Schneider, Zaremba, and
  Abbeel]{DR}
Josh Tobin, Rachel Fong, Alex Ray, Jonas Schneider, Wojciech Zaremba, and
  Pieter Abbeel.
\newblock Domain randomization for transferring deep neural networks from
  simulation to the real world.
\newblock In \emph{2017 IEEE/RSJ international conference on intelligent robots
  and systems (IROS)}, pp.\  23--30. IEEE, 2017.

\bibitem[White(2016)]{white2016sampling}
Tom White.
\newblock Sampling generative networks.
\newblock \emph{arXiv preprint arXiv:1609.04468}, 2016.

\bibitem[Wilson et~al.(2021)Wilson, Qi, Agarwal, Lambert, Singh, Khandelwal,
  Pan, Kumar, Hartnett, Kaesemodel~Pontes, Ramanan, Carr, and Hays]{argoverse2}
Benjamin Wilson, William Qi, Tanmay Agarwal, John Lambert, Jagjeet Singh,
  Siddhesh Khandelwal, Bowen Pan, Ratnesh Kumar, Andrew Hartnett, Jhony
  Kaesemodel~Pontes, Deva Ramanan, Peter Carr, and James Hays.
\newblock Argoverse 2: Next generation datasets for self-driving perception and
  forecasting.
\newblock In J.~Vanschoren and S.~Yeung (eds.), \emph{Proceedings of the Neural
  Information Processing Systems Track on Datasets and Benchmarks}, volume~1,
  2021.
\newblock URL
  \url{https://datasets-benchmarks-proceedings.neurips.cc/paper/2021/file/4734ba6f3de83d861c3176a6273cac6d-Paper-round2.pdf}.

\bibitem[Xu \& Raginsky(2017)Xu and Raginsky]{MI_gengap_SL}
Aolin Xu and Maxim Raginsky.
\newblock Information-theoretic analysis of generalization capability of
  learning algorithms.
\newblock \emph{Advances in Neural Information Processing Systems}, 30, 2017.

\bibitem[Xu et~al.(2019)Xu, Hu, Leskovec, and Jegelka]{GIN}
Keyulu Xu, Weihua Hu, Jure Leskovec, and Stefanie Jegelka.
\newblock How powerful are graph neural networks?
\newblock In \emph{International Conference on Learning Representations}, 2019.
\newblock URL \url{https://openreview.net/forum?id=ryGs6iA5Km}.

\bibitem[Zhang et~al.(2020)Zhang, Abbeel, and Pinto]{zhang2020automatic}
Yunzhi Zhang, Pieter Abbeel, and Lerrel Pinto.
\newblock Automatic curriculum learning through value disagreement.
\newblock \emph{Advances in Neural Information Processing Systems},
  33:\penalty0 7648--7659, 2020.

\end{thebibliography}
\bibliographystyle{iclr2024_conference}

\newpage

\appendix

\section{theoretical results}\label{app:proofs}

\begin{lemma}
    (proof in appendix) Given a set of training levels $L$ and an agent model $\pi = f \circ b$, where $b(H^o_t) = h_t$ is an intermediate state representation and $f$ is the policy head, we can bound $\mut (L,\pi \circ b)$ by $\mut (L,b)$, which in turn satisfies
    \begin{align}
        \mut (L,\pi) \leq \mut (L,b) &= \mathcal{H}(p(\ri)) + \sum_{i\in L} \int dh p(\rh, \ri) \log p(\ri|\rh) \\
        &\approx \mathcal{H}(p(\ri)) + \frac{1}{|B|} \sum_{(i, H_t^o) \in B} \log p(\ri|b(H_t^o))
    \end{align}
    where $\mathcal{H}(p)$ is the entropy of distribution $p$ and $B$ is a batch of trajectories collected from individual levels $i \in L$.

    proof:
    
    Given that the information chain of our model follows $H_t^o \rightarrow b \rightarrow f$, we have $\mut (L,f \circ b) \leq \mut (L,b)$ following the data processing inequality. $\mut (L,b)$ can then be manipulated as follows
    \begin{align}\label{eq:proof_mi_kld}
        \mut(L,b) &= \sum_{i \in L} \int d\vh p(\rvh,i) \log \frac{p(\rvh,i)}{p(\rvh)p(\ri)} \\
        &= - \sum_{i \in L} \int d\rvh p(\rvh,\ri) \log p(i) + \sum_{i \in L} \int d\vh p(\rvh, i) \log p(\ri|\rvh) \\
        &\approx \mathcal{H}(p(\ri)) + \frac{1}{|B|} \sum^{|B|}_{n} \log p(i^{(n)}|\vh^{(n)}) \label{eq:proof_empirical_approx}
    \end{align}

    where in \cref{eq:proof_empirical_approx} we approximate $p(\vh,i)$ as the empirical distribution

    \begin{equation} \label{eq:empirical_approx}
        \tilde{p}(\vh,i) = 
        \begin{cases}
            \frac{1}{|B|} &\text{ if } (H_t^o, i) \in B, \text{ with } \vh = b(H_t^o) \\
            0 &\text{ otherwise.}
        \end{cases}
    \end{equation}

    Note that we can treat non-recurrent architectures as a particular case, setting $H^o_t = o_t$ without loss of generality.
    
\end{lemma}

\section{additional experimental results}\label{app:results}

\subsection{procgen additional experimental results}\label{app:procgen_results}

\begin{figure}[!htb]
\centering
    \begin{subfigure}{0.49\linewidth}
            \includegraphics[width=1\linewidth]{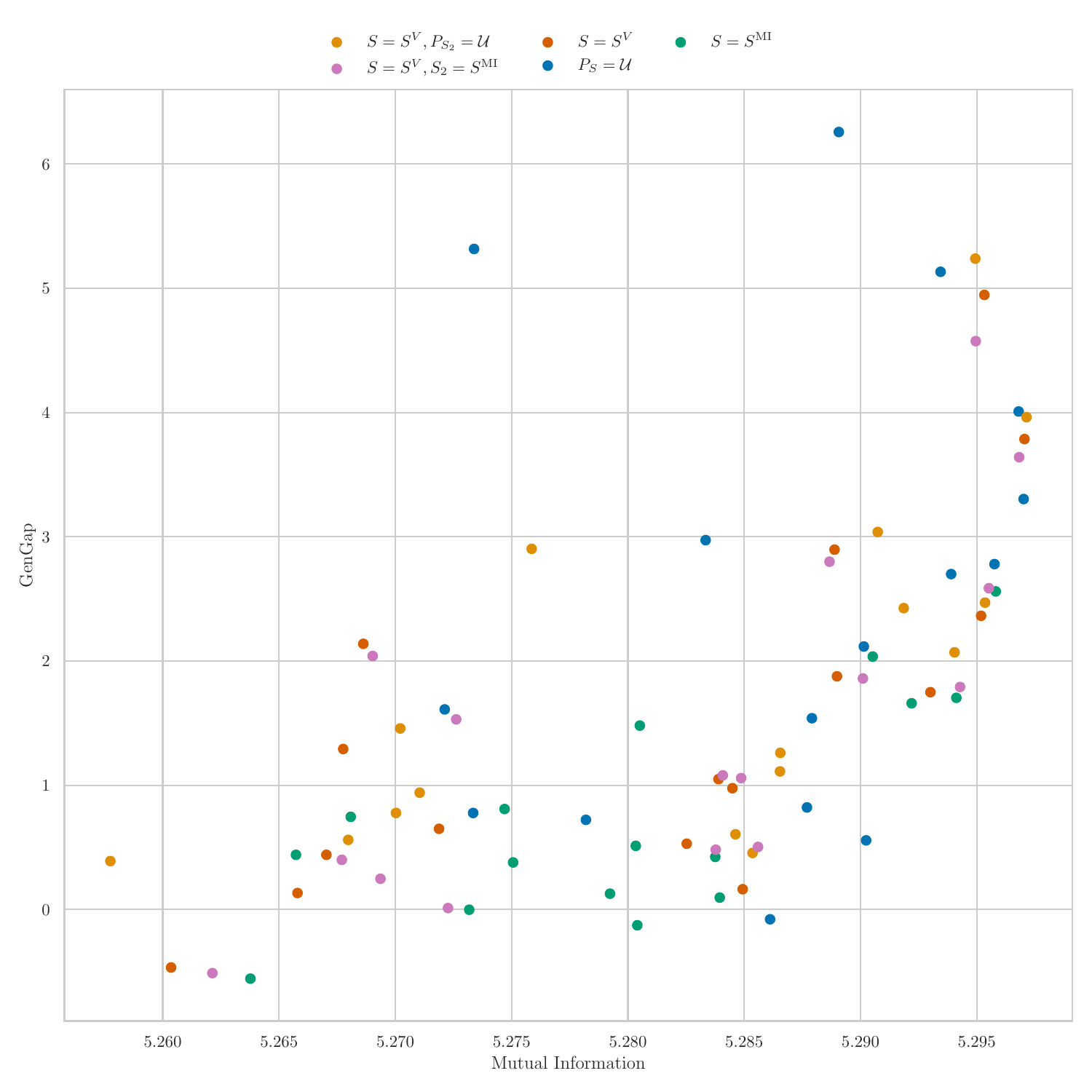}
    \end{subfigure}%
    \begin{subfigure}{0.49\linewidth}
            \includegraphics[width=1\linewidth]{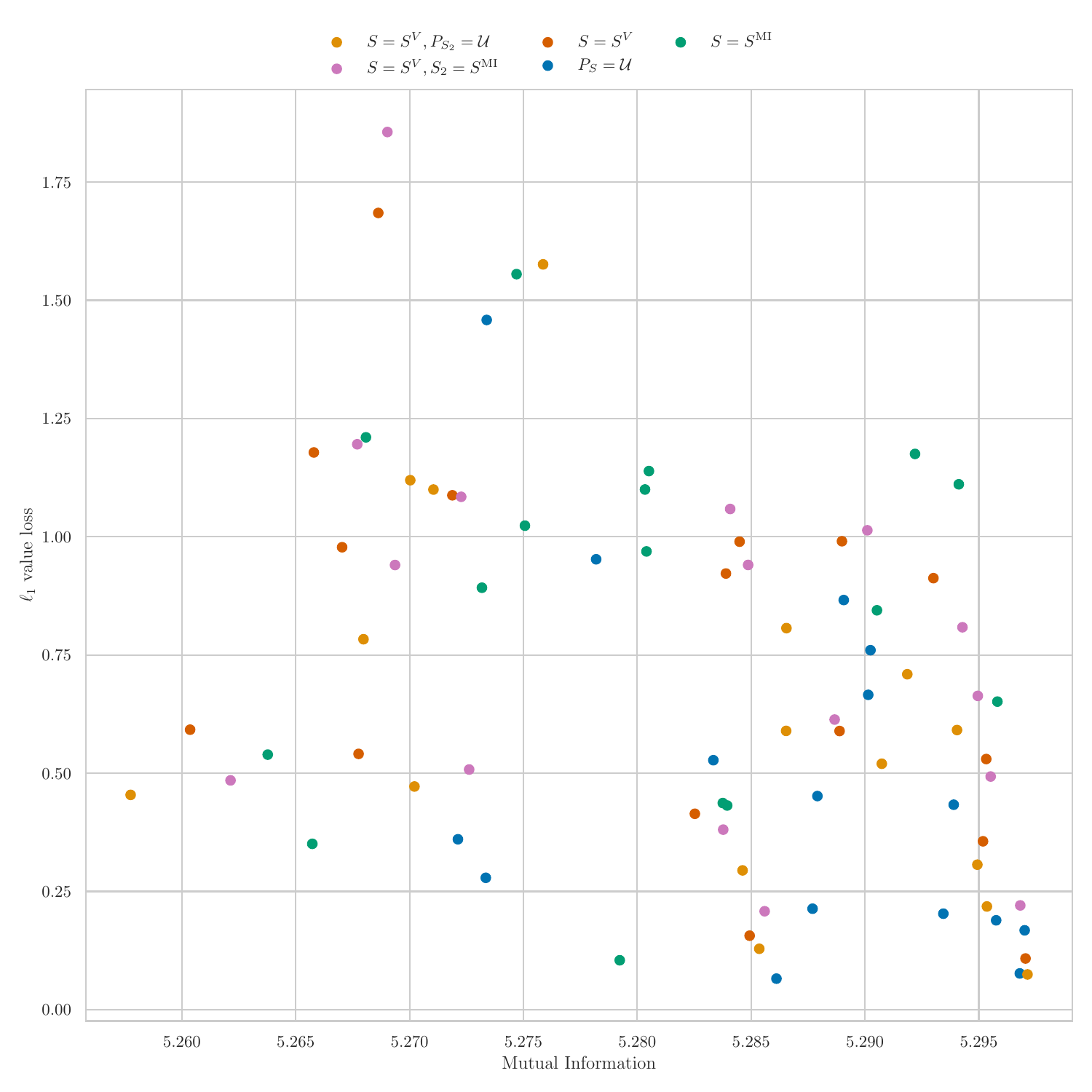}
    \end{subfigure}%
    \caption{Scatter plot displaying the relationship between $\mut(L,b)$ and the (unormalised) $\text{GenGap}$ (left) and with the $\ell_1$ average value loss (right), measured across all methods and Procgen games at the end of training. Each point represents 5 seeds of a level sampling method in a particular game.}\label{fig:correlation_analysis}
\end{figure}

\begin{figure}[!htb]
\centering
    \begin{subfigure}{0.49\linewidth}
            \includegraphics[width=1\linewidth]{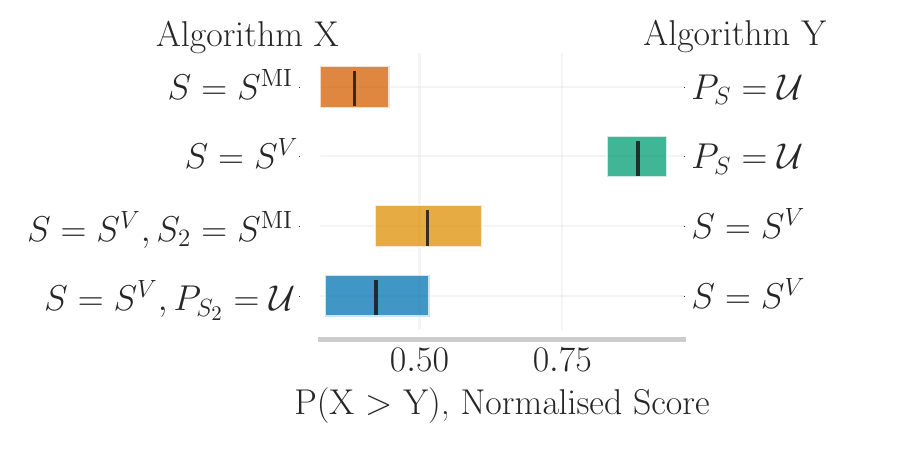}
    \end{subfigure}%
    \begin{subfigure}{0.49\linewidth}
            \includegraphics[width=1\linewidth]{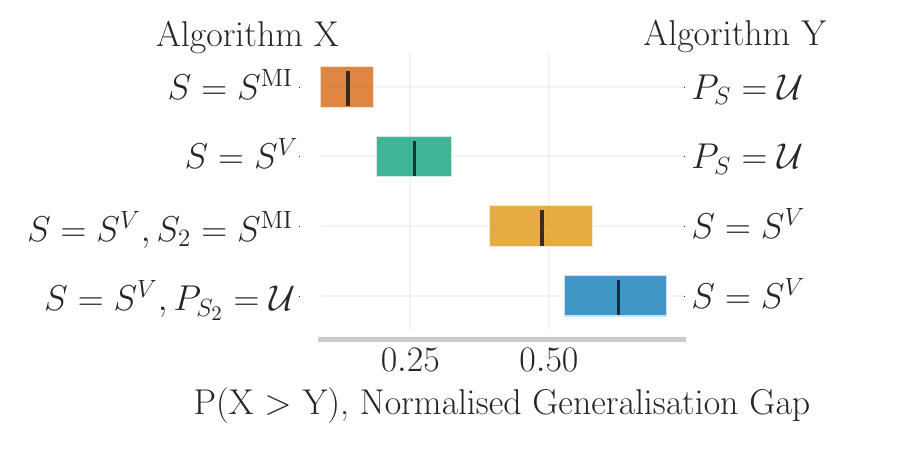}
    \end{subfigure}%
    \caption{Probability of algorithm $X$ incurring a higher normalised test score (left) and $\text{GenGap}$ (right) than algorithm $Y$. Evaluation performed over 5 seeds across all Procgen games, using the rliable library\citep{rliable}. Colored bars indicate the 95\% confidence interval.}\label{fig:p_improvement_procgen}
\end{figure}

From \Cref{eq:empirical_approx2}, we estimate $\mut(L, b)$ modelling $p_\vtheta$ as a linear classifier. We plot this estimate against the $\text{GenGap}$ and the $\ell_1$ value loss for all methods tested and across Procgen games in \Cref{fig:correlation_analysis}. As expected under our theoretical framework, we measure a positive correlation between $\mut(L, b)$ and $\text{GenGap}$ with Kendall rank correlation coefficient $\tau=0.53$ ($p<1\mathrm{e}{-34}$), and a negative correlation with the $\ell_1$ value loss with Kendall rank correlation coefficient $\tau=-0.28$ ($p<1\mathrm{e}{-16}$).

In order to provide a more intuitive quantification of our mutual information estimates, we consider the classification accuracy of the linear classifier used to compute our estimate for $\mut(L,b)$, as these two quantities are proportional with each other. Out of 200 training levels, the classifier correctly predicts the current level $49\%$ of the times under uniform sampling, $34\%$ under $(S=S^V)$ and $23 \%$ under $S^\mut$. Adaptive sampling strategies are therefore able to reduce $(S=S^\mut)$ across ProcGen games, and ranking different methods according to their level classification accuracy will also sort them according to their respective $\text{GenGap}$. To understand how likely it is for a given sampling strategy to improve over another, we report the probability of improvement in test scores and $\text{GenGap}$ for different pairs of strategies in \Cref{fig:p_improvement_procgen}.

Nevertheless, the mean classifier accuracy remains $68$ times random guessing for $(S=S^V)$ and $46$ times random guessing for $(S=S^\mut)$. As the classifier makes a prediction using an internal representation obtained from a single observation we find these results surprising, and demonstrate adaptive sampling strategies can only reduce $\mut(L,b)$ up to a point. To further reduce $\mut(L,b)$, adaptive sampling should be combined with other data regularisation techniques, such as the level augmentation technique proposed by SSED and/or additional data augmentation techniques. Indeed \cite{PLR} report a significant improvement in test scores when combining PLR with UCB-DrAC \cite{raileanu2021UCB-DrAC}, an observation augmentation method. 

\subsection{comparing the effectiveness of adaptive sampling strategies across procgen games}\label{app:procgen_level_analysis}

\begin{figure}[!htb]
\centering
    \begin{subfigure}{0.4\linewidth}
            \includegraphics[width=1\linewidth]{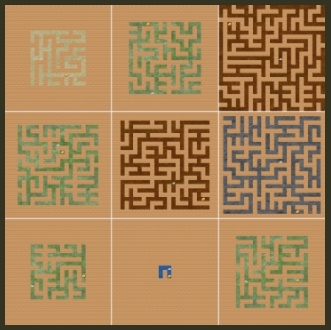}
    \end{subfigure}%
    \begin{subfigure}{0.4\linewidth}
        \includegraphics[width=1\linewidth]{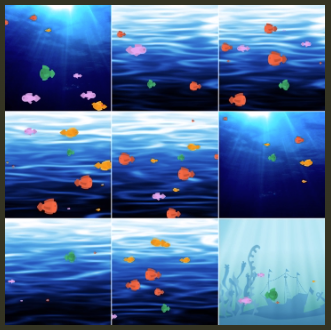}
    \end{subfigure}%
    \caption{Agent observations sampled from 9 levels from the Maze (left) and Bigfish (right) games of the Procgen benchmark.}\label{fig:procgen_levels}
\end{figure}

We observe that both the classifier accuracy under uniform level sampling and the potential improvement induced by adaptive sampling is highly dependent on the procgen game tested. To better understand why, we compare the measured accuracy with a qualitative analysis of the observations and levels encountered in the Maze and Bigfish games, which we provide a sample of in \Cref{fig:procgen_levels}.

In Maze, the accuracy remains over $80\%$ ($160\times$ random) for all methods tested and the reduction is $\text{GenGap}$ insignificant. On the other hand, in Bigfish all adaptive sampling strategies tested lead to a significant reduction in classifier accuracy, dropping from $80\%$ to under $20\%$, and they are associated with a significant drop in $\text{GenGap}$ and improvement in test scores. 

In Maze, the observation space is set up such that the agent observes the full layout at each timestep. The layout is unique to each level and provides many features for identification that are straightforward to learn by the agent's ResNet architecture. In addition, these features cannot be ignored by the agent model in order to solve the task. Intuitively, we can hypothesise that adaptive sampling strategies will not be effective if all the levels are easily identifiable by the agent, which appears to be the case in Maze. In these cases, other data regularisation techniques, such as augmenting the observations, can be more effective, and in fact \cite{PLR} report that Maze is one of the games where combining PLR with UCB-DrAC leads to a significant improvement in test scores.

On the other hand, we observe that many of the Bigfish levels yield similar observations. Indeed, both the features relevant to the task (the fish) and irrelevant (the background) are similar in many of the training levels. Furthermore, there's significant variation in the observations encountered during an episode, as fish constantly appear and leave the screen. Yet, some levels (top left, middle and bottom right) are easily identifiable thanks to their background, and we can hypothesise that adaptive sampling strategies will tend to de-prioritise them more often, essentially performing data regularisation via a form of rejection sampling.

\subsection{quantifying the distributional shift in minigrid}\label{app:dist-shift}

\begin{figure}[!htb]
    \centering
    \begin{subfigure}{0.33\linewidth}
            \includegraphics[width=1\linewidth]{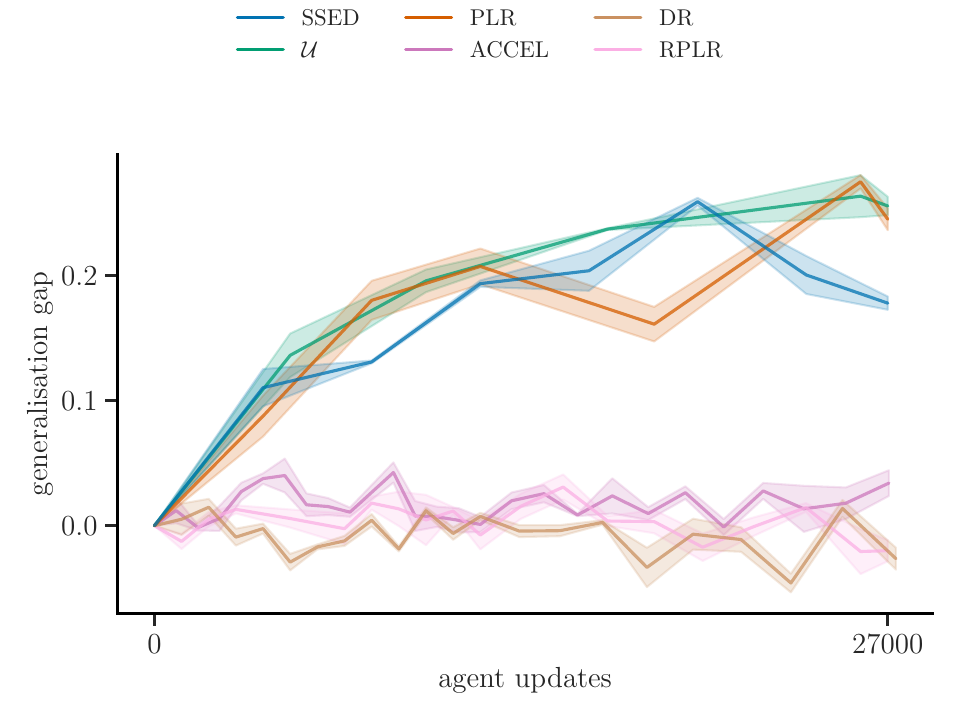}
            \caption{}\label{subfig:gengap}
    \end{subfigure}%
        \begin{subfigure}{0.33\linewidth}
            \includegraphics[width=1\linewidth]{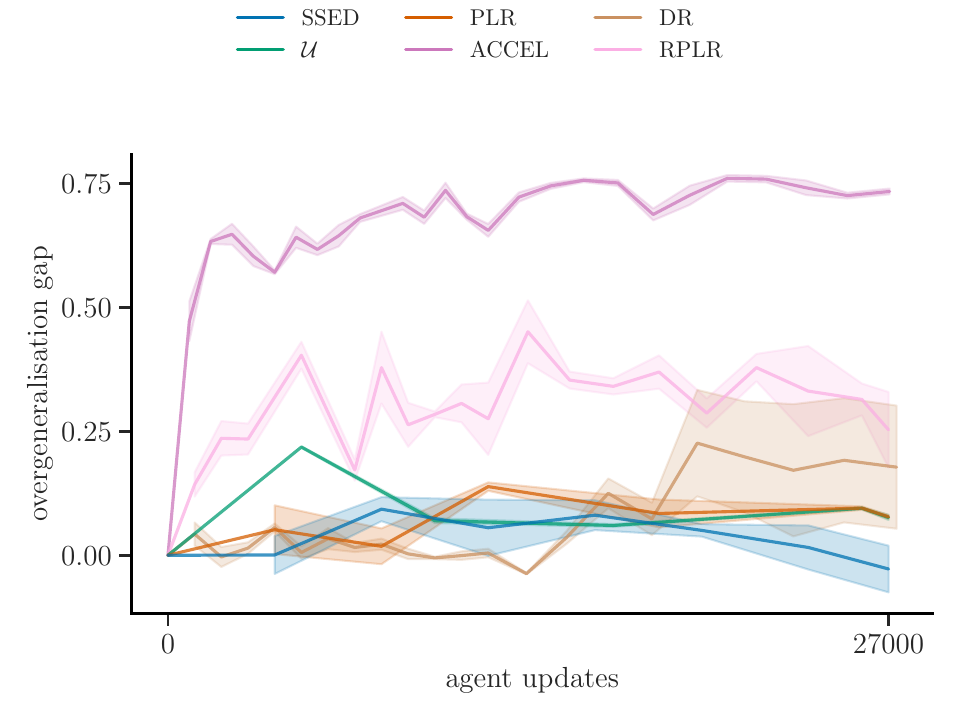}
            \caption{}\label{subfig:overgengap}
    \end{subfigure}%
        \begin{subfigure}{0.33\linewidth}
            \includegraphics[width=1\linewidth]{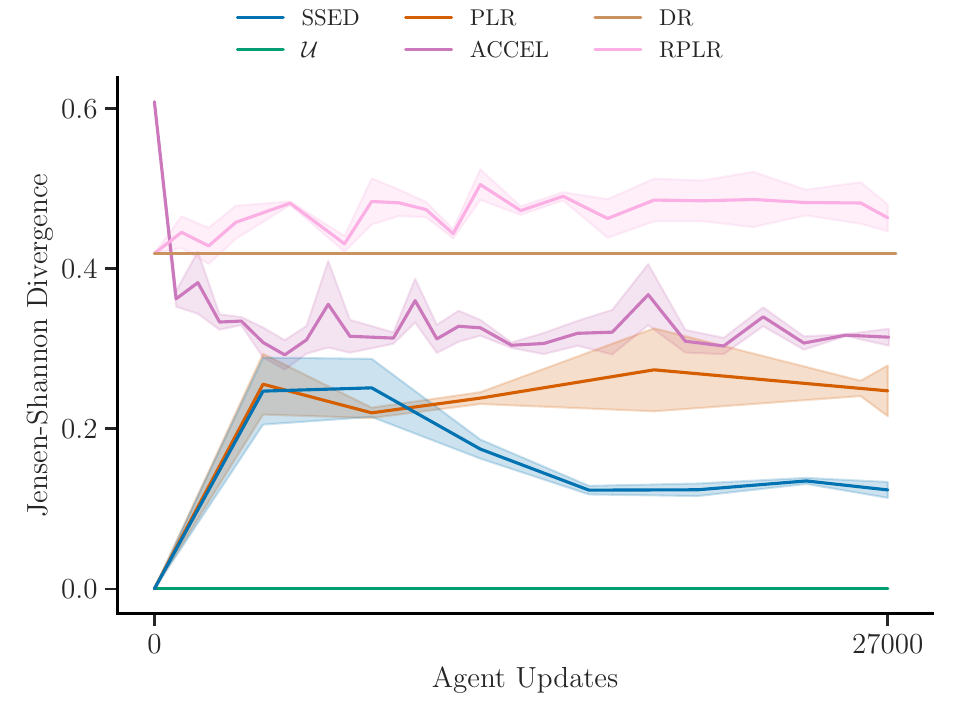}
            \caption{}\label{subfig:jsd}
    \end{subfigure}%
    \caption{Generalisation gap ((a), \Cref{eq:GenGap}) and over-generalisation gap ((b), \Cref{eq:OverGenGap}) of different methods during training. Fixed set sampling strategies experience higher generalisation gap, while UED methods are dominated by the over-generalisation gap. SSED tends to follow a similar profile as fixed set sampling methods, with a moderate generalisation gap and a low (and even at times negative) over-generalisation gap. SSED also exhibit small distributional shift in its level parameters, as demonstrated by the evolution of the JSD over the course of training (c). Surprisingly, SSED demonstrates a smaller divergence than PLR, even when PLR only has access to $X_\text{train}$ and as such can only affect the JSD by changing the prioritisation of individual levels $i_\vx$, $\vx \in X_\text{train}$.
    }
    \label{fig:gen_gaps}
\end{figure}

In \Cref{fig:gen_gaps}, we report the generalisation and over-generalisation gaps in the Minigrid experiments. We observe that UED methods tend to exhibit lower generalisation gaps than SSED, PLR or uniform sampling. We find that introducing an additional metric in the form of the $\text{OverGap}$  \Cref{eq:OverGenGap} necessary to quantify this form of misgeneralisation. We next study the correlation between the $\text{OverGap}$ and distributional shifts between the underlying CMDP level parameter distribution $p(\rvx)$ and $p_\Lambda(\rvx)$, the distribution of level parameters existing in the level replay buffer $\Lambda$.

We approximate $p(\rvx)$ as $\tilde{p}(\rvx) \approx \mathcal{U}(X_\text{train})$ and we use the Jensen-Shannon Divergence (JSD) between $\tilde{p}(\rvx)$ and $p_\Lambda(\rvx)$, defining the JSD in a space consistent with the CMDP semantics. To do so, we measure the distribution of distances between the goal location and each other tile type. The JSD is therefore expressed as $\text{JSD}(c_p||c_q)$, where $c(t,d|\rvx)$ is the categorical distribution measuring the probability of tile type $t$ occurring at distance $d$ from the goal location in a given level $\vx$ and $c_p$ is the marginal $c_p = \mathbb{E}_{\rvx \sim p(\rvx)}[c(t,d|\rvx)]$, setting $p_\Lambda(\rvx)$ and $q=\mathcal{U}(X_\text{train})$.

\begin{figure}[!htb]
    \centering
        \begin{subfigure}{0.33\linewidth}
            \includegraphics[width=1\linewidth]{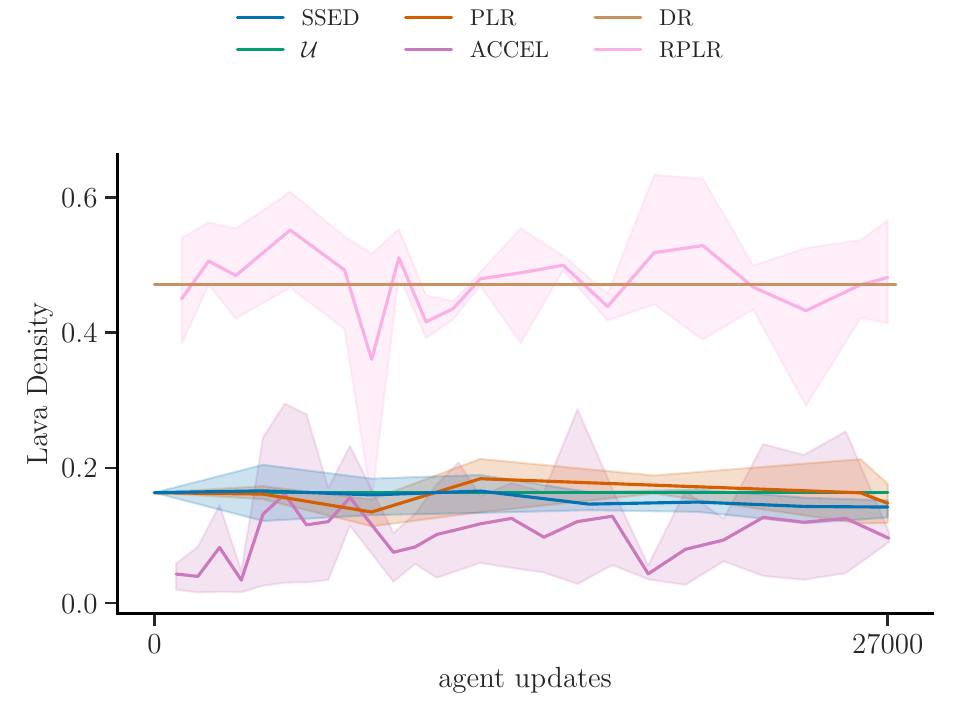}
    \end{subfigure}%
            \begin{subfigure}{0.33\linewidth}
            \includegraphics[width=1\linewidth]{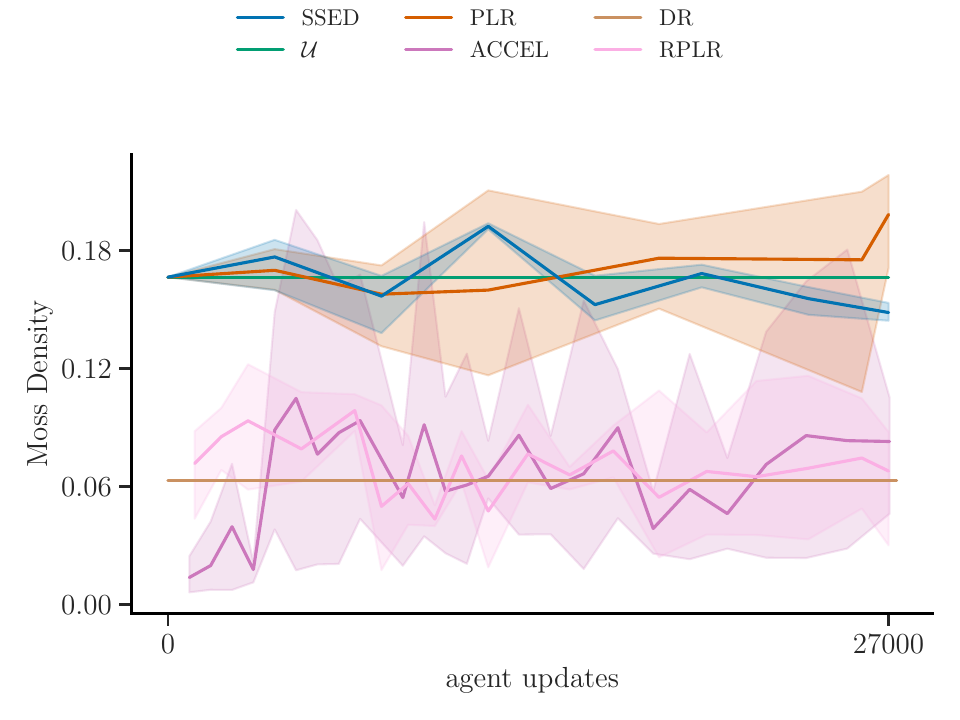}
    \end{subfigure}%
    \begin{subfigure}{0.33\linewidth}
            \includegraphics[width=1\linewidth]{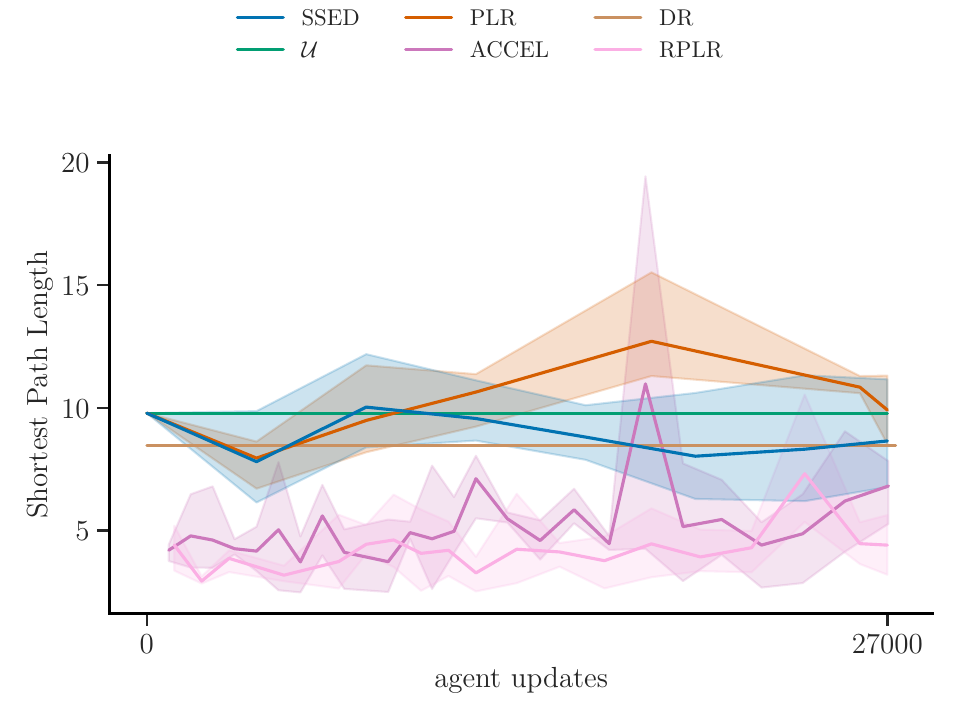}
    \end{subfigure}%
    \caption{Left and middle: evolution of lava and moss tile densities encountered in sampled levels over the course of training. Right: evolution of the shortest path length between the start and goal location in sampled levels over the course of training.
    }
    \label{fig:buffermetrics}
\end{figure}

We report how the JSD evolves over the course of training for different methods in \Cref{subfig:jsd}. We observe that distributional shift occurs early on during training and remains relatively stable afterwards in all methods. JSD and OverGap tend to be positively correlated for most methods, except for DR and PLR, which both present high JSD but low OverGap. SSED is the only generative method to maintain a low JSD throughout training%
In \cref{fig:buffermetrics}, we report additional metrics on the levels sampled by each method. We find that SSED tends to be as proficient as PLR in maintaining consistency with $X_\text{train}$ with the occurence of different tile types, as well as higher order task-relevant properties such as the shortest path length between the start and goal locations.

\section{cmdp specification and the level generation process}\label{app:cmdp_levelset}

In Minigrid \cite{gym_minigrid}, the agent receives as an observation a partial view of its surroundings (in our experiments it is set to two tiles to each side of the agent and four tiles in front) and a one-hot vector representing the agent's heading. The action space consists of 7 discrete actions, however in our setting only the actions moving the agent forward and rotating it to the left or right have an effect on the environment. The episode starts with the agent at the start tile and facing a random direction. The episode terminates successfully when the agent reaches the goal tile and receives a reward between 0 and 1 based on the number of timesteps it took to get there. The episode will terminate without a reward if the agent steps on a lava tile, or when the maximum number of timesteps is reached.

\begin{figure}[!htb]
    \centering
            \includegraphics[width=.4\linewidth]{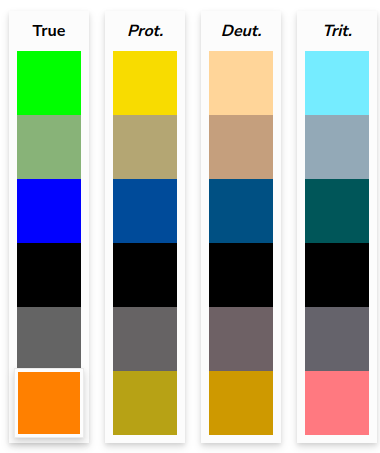}
    \caption{Color palette used for rendering minigrid layouts in this paper and their equivalent for Protanopia (Prot.), Deuteranopia (Deut.) and Tritanopia (Trit.) color blindness. We refer to each row in the main text as, in order: green (goal tiles), pale green (moss tiles), blue (agent), black (empty/floor tiles), grey (wall tiles) and orange (lava tiles).
    }
    \label{fig:color_palette}
\end{figure}

Levels are parameterised as 2D grids representing the overall layout, with each tile type represented by an unique ID. Tiles can be classified as navigable (for example, moss or empty tiles) or non-navigable (for example, walls and lava, as stepping into lava terminates the episode). To be valid, a level must possess exactly one goal and start tile, and to be solvable there must exist a navigable path between the start and the goal location. We provide the color palette of tiles used in \Cref{fig:color_palette}.

\begin{figure}[!htb]

\sbox\twosubbox{%
  \resizebox{\dimexpr.9\textwidth-1em}{!}{%
    \includegraphics[height=3cm]{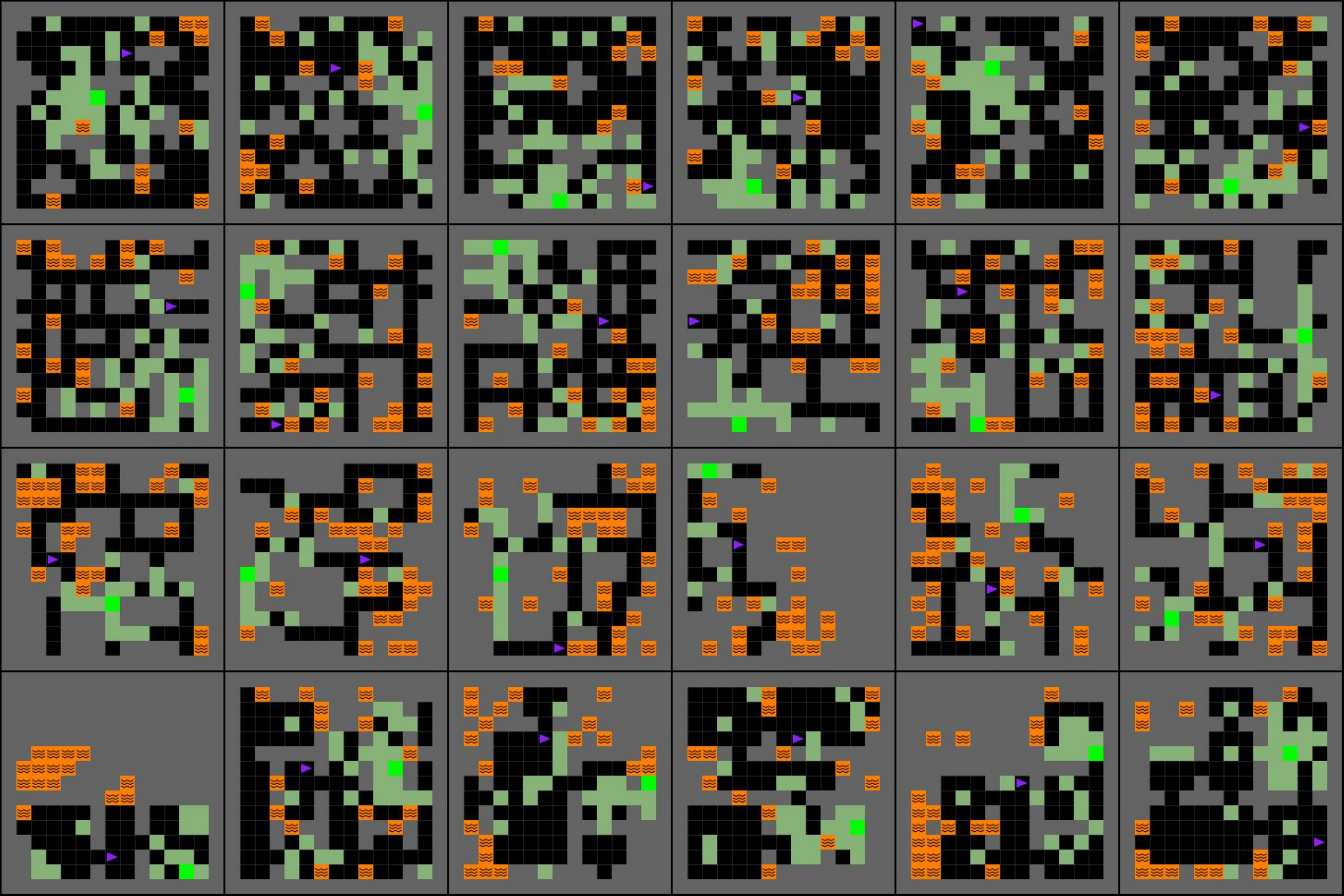}%
    \includegraphics[height=3cm]{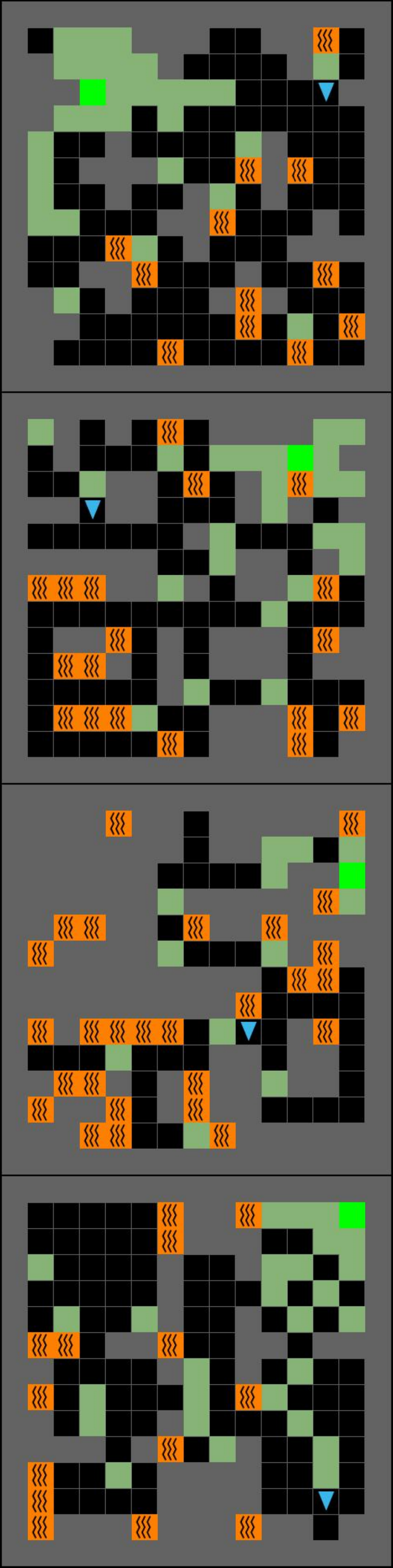}%
  }%
}
\setlength{\twosubht}{\ht\twosubbox}

\centering

\subcaptionbox{Example training levels from $X_\text{train}$\label{f}}{%
  \includegraphics[height=\twosubht]{app_content/layouts/cave_escape_dataset.pdf}%
}\quad
\subcaptionbox{Held-out levels\label{s1}}{%
  \includegraphics[height=\twosubht]{app_content/layouts/cave_escape_16k_4ts_id_rotated.pdf}%
}

\caption{Sample levels from $X_\text{train}$ and from held-out test levels. Wall tiles are rendered in gray, empty tiles in black, moss tiles in green and the goal tile in lime green. The agent is rendered as a blue or purple triangle, and is depicted at its start location. Each row corresponds to levels generated with a specific wave function collapse base pattern. Four different base patterns are used in $X_\text{train}$.}\label{fig:layouts_dataset}

\end{figure}

In this work, we define and train within the ``Cave Escape'' CMDP, which corresponds to a subset of the solvable levels in which moss and lava node placement is respectively positively and negatively correlated with the geodesic distance to the goal.\footnote{To measure the distance-to-goal of a non-navigable node, we first find the navigable node that is closest from it and measure its geodesic distance to the goal. We then add to it the distance between this navigable node and the non-navigable node of interest. If there are multiple equally close navigable nodes, we select the navigable node with the smallest geodesic distance to the goal. %
} Under partial observability, the minimax regret policy for this CMDP would leverage moss and lava locations as context cues, seeking regions with perceived higher moss density avoiding regions with perceived high lava density, and thus the CMDP possesses an attributable goal and optimal behavior. We provide example levels of the CMDP in \Cref{fig:layouts_dataset}.

\begin{figure}[!htb]

\sbox\twosubbox{%
  \resizebox{\dimexpr.9\textwidth-1em}{!}{%
    \includegraphics[height=3cm]{app_content/layouts/cave_escape_dataset.pdf}%
    \includegraphics[height=3cm]{app_content/layouts/cave_escape_16k_4ts_id_rotated.pdf}%
  }%
}
\setlength{\twosubht}{\ht\twosubbox}

\centering

\subcaptionbox{Example levels from the first set of edge cases}{%
  \includegraphics[height=\twosubht]{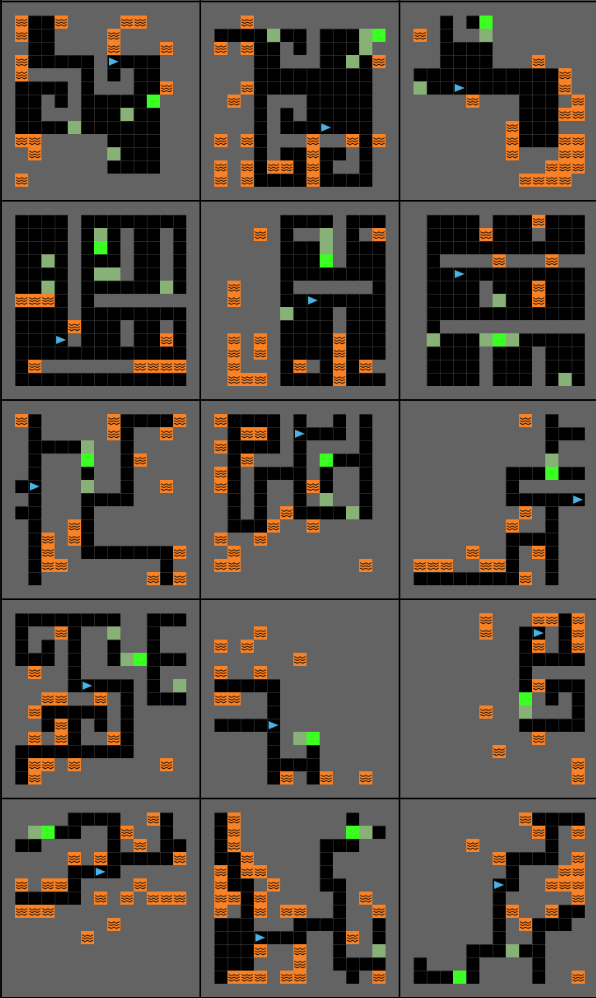}%
}\quad
\subcaptionbox{Example levels from the second set of edge cases\label{s2}}{%
  \includegraphics[height=\twosubht]{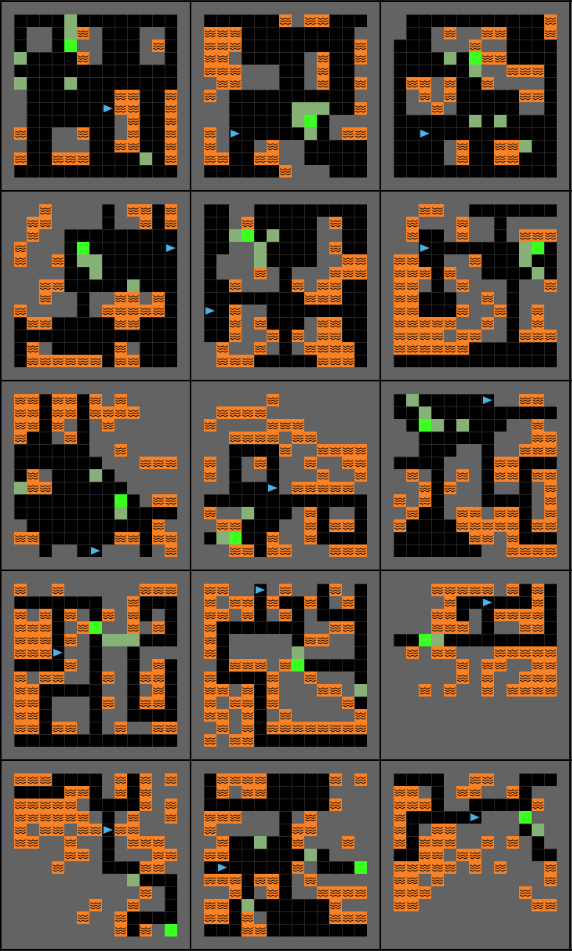}%
}

\caption{We generate 2 separate sets of edge cases, using 14 different base patterns not used to generate $X_\text{train}$. The moss density of the first set (a) is 3 times as small as in $X_\text{train}$, making finding the goal using CMDP contextual cues more challenging. (b) is the same as (a) but also has the lava density multiplied by a factor of 3, making it more difficult to avoid walking into lava and failing the episode. Both sets are combined when evaluating the agent on edge cases.}\label{fig:layout_edge_cases}

\end{figure}

\subsection{Generating Highly Structured Levels}

We employ the wave function collapse (WFC) algorithm \citep{WFC} as our procedural generation algorithm to obtain highly structured but still diverse gridworld layouts. WFC takes as an input a basic pattern and gradually collapses a superposition of all possible level parameters into a layout respecting the constraints defined by the input pattern. By doing so, it is possible to generate a vast number of tasks from a small number of starting patterns. Given suitable base patterns the obtained layouts provide a high degree of structure, and guarantee that both task structure and diversity scale with the gridworld dimensions. We provide 22 %
different base patterns and allow for custom ones to be defined. After generating a layout using WFC, we convert the navigable nodes of a layout into a graph, choose its largest connected component as the layout and convert any unreachable nodes to non-navigable nodes. We place the goal location at random and place the start at a node located at the median geodesic distance from the goal in the navigation graph. By doing so we ensure that the complexity of generated layouts is relatively consistent given a specific grid size and base pattern. Finally we sample tiles according to parameterisable distributions defined over the navigable or non-navigable node sets. In this work the tile set consists of the $\{$ moss, empty, start, goal $\}$ tiles as the navigable set and the $\{$ wall, lava$\}$ tiles as the non-navigable set, and we define distributions for the moss and lava node types over the navigable and non-navigable node sets, respectively.

\subsection{Controlling Level Complexity}

We provide two options to vary the complexity of the level distribution. The first is to change the gridworld size, which directly results in an increase in complexity. The second, which is specific to the Cave Escape CMDP, is to change the sampling probability of moss and lava nodes. Since the environment is partially observable, reducing the fraction of moss to navigable nodes, or increasing their entropy diminishes the usefulness of moss tiles as context cues. On the other hand, increasing the density of lava tiles increases the risk associated with selecting the wrong action during play. Thus it is straightforward assess the agent's performance on edge-cases by defining level sets with a larger layout size, or with shifted moss and lava tile distributions.  

\section{implementation details}\label{app:imp_details}

\subsection{procgen}\label{app:procgen}
The Procgen Benchmark is a set of 16 diverse PCG environments that echo the gameplay variety seen in the ALE benchmark \cite{ALE}. The game levels, determined by a random seed, can differ in visual design, navigational structure, and the starting locations of entities. All Procgen environments use a common discrete 15-dimensional action space and generate $64 \times 64 \times 3$ RGB observations. A detailed explanation of each of the 16 environments is given by \cite{procgen}. Leading RL algorithms such as PPO reveal significant differences between test and training performance in all games, making Procgen a valuable tool for evaluating generalisation performance. 

We conduct our experiment on the easy setting of Procgen, which employs 200 training levels and a budget of 25M training steps, and evaluate the agent's ZSG performance on the full range of levels, excluding the training levels. We calculate normalised test returns using the formula $\frac{(R-R_\text{min})}{(R_\text{max} - R_\text{min})}$, where $R$ is the non normalised return and $R_\text{min}$ and $R_\text{max}$ are the minimum and maximum returns for each game as provided in \citep{procgen}.

We employ the same ResNet policy architecture and PPO hyperparameters used across all games as \cite{procgen}, which we reference in \Cref{table:hyperparams}. To compute the MI based scoring strategy $S^\text{MI}$ used in our experiments, we parametrise $p_\vtheta(\lvl|b(o_t)$ as a linear classifier and we ensure the training processes of the agent and the classifier remain independent from one-another by employing a separate optimiser and stopping the gradients from propagating through the agent's network.

\subsection{minigrid rl agent}

We use the same PPO-based agent as reported in \cite{ACCEL}. The actor and critic share the same initial layers. The first initial layer consists of a convolutional layer with 16 output channels and kernel size 3 processes the agent's view and a fully connected layer that processes its directional information. Their output is concatenated and fed to an LSTM layer with hidden size 256. The actor and critic heads each consist of two fully connected layers of size 32, the actor outputs a categorical distribution over action probabilities while the critic outputs a scalar. Weights are optimized using Adam and we employ the same hyperparameters in all experiments, reported in \Cref{table:hyperparams}. Trajectories are collected via 36 worker threads, with each experiment conducted using a single GPU and 10 CPUs.

Following \cite{ACCEL}, the non dataset based methods employ domain randomisation as their standard level generation process, in which the start and goal locations, alongside a random number between 0 and 60 moss, wall or lava tiles are randomly placed. The level editing process of ACCEL and SSED-EL remains unchanged from \cite{ACCEL}, consisting of five steps. The first three steps may change a randomly selected tile to any of its counterparts, whereas the last two are reserved to replacing the start and goal locations if they had been removed in prior steps.

We train three different seeds for each baseline. We use the same hyperparameters as reported in \cite{ACCEL} for the DR, RPLR and ACCEL methods and as \cite{PLR} for PLR, as an extensive hyperparameter search was conducted in a similar-sized Minigrid environment in each case. SSED employs the same hyperparameters as PLR for its level buffer, with the additional secondary sampling strategy hyperparameters introduced by SSED. We did not perform an hyperparameter search for these additional hyperparameters as we found that the initial values worked adequately. We report all hyperparameters in \Cref{table:hyperparams}.

\subsection{vae architecture and pre-training process}\label{app:vae_imp}

We employ the $\beta-$VAE formulation proposed in \cite{betaVAE}, and we parametrise the encoder as a Graph Convolutional Network (GCN), a generalisation of the Convolutional Neural Network (CNN) \citep{CNNimagenet} to non Euclidian spaces. Our choice of a GCN architecture is motivated by the fact that the level parameter space $\sX$ is simulator-specific. Employing a graph as an input modality for our encoder gives our model additional flexibility to by applicable to different simulators. Using a GCN, some of the inductive biases that would be internal in a traditional architecture can be defined through the wrapper encoding the environment parameter $\vx$ into the graph $\mathcal{G}_\vx$. For example, in minigrid, we represent each layout as a grid graph, each gridworld cell being an individual node. Doing so makes our GCN equivalent to a traditional CNN in the Minigrid domain. We select the GIN architecture \citep{GIN} for the GCN, which we connect to an MLP network that outputs latent distribution parameters $\vmu_\vz, \vsigma_vz$. The decoder is a fully connected network with three heads. The \textit{layout} head outputs the parameters of Categorical distributions for each grid cell, predicting the tile identity between [Empty, Moss, Lava, Wall]. The \textit{start} and \textit{goal} heads output the parameters of Categorical distributions predicting the identity of the start and goal locations across grid cells, which matches the inductive bias of a single goal or start node being present in any given level.

We pre-train the VAE for 200 epochs on $X_\text{train}$, using cross-validation for hyperparameter tuning. During training, we formulate the reconstruction loss as a weighted sum of the cross-entropy loss for each head.\footnote{To compute the cross-entropy loss of the layout head, we replace the start and goal nodes in the reconstruction targets by a uniform distribution across \{moss, empty\}.} At deployment, we guarantee \textit{valid} layouts (layouts containing a unique start and goal location, but not necessarily solvable) by masking the non-passable nodes sampled by the layout head when sampling the start location, and masking the generated start and non-passable nodes when sampling the goal location. In this way, we guarantee unique start and goal locations that will not override one-another. Note that our generative model may still generate \textit{unsolvable} layouts, which do not have a passable path between start and goal locations, and therefore it must learn to generate solvable layouts from $X_\text{train}$ in order to be useful. We do not explicitly encourage the VAE to generate solvable layouts, but we find that optimising for the ELBO in \Cref{eq:ELBO} is an effective proxy. Layouts reconstructed from $X_\text{train}$ have over 80\% solvability rate, while layouts generated via latent space interpolations have over 70\% solvability rate. This indicate that maximising the ELBO results in the generative model learning to reproduce the high-level abstract properties that are shared across the level parameters in the training set, which we also observe in \Cref{fig:buffermetrics}.

We conduct a random sweep over a budget of 100 runs, jointly sweeping architectural parameters (number of layers, layer sizes) and the $\beta$ coefficient, individual decoder head reconstruction coefficients and the learning rate. Each run takes 20 minutes, which means that our sweeping procedure takes less time to complete than a single seed of our Minigrid experiment. We report the chosen hyperparameters in \Cref{table:hyperparams_vae}.

\begin{table}[!htb]
\caption{Hyperparameters used for Minigrid experiments. Hyperparameters shared between methods are only reported if they change from the method above.}
\label{table:hyperparams}
\begin{center}
\scalebox{0.88}{
\begin{tabular}{lcccc}
\toprule
\textbf{Parameter} & Procgen & MiniGrid\\
\midrule
\emph{PPO} & \\
$\gamma$ &0.999 & 0.995 \\
$\lambda_{\text{GAE}}$ & 0.95 & 0.95  \\
PPO rollout length & 256 & 256 \\
PPO epochs & 3 & 5 \\
PPO minibatches per epoch & 8 & 1 \\
PPO clip range & 0.2 & 0.2 \\
PPO number of workers & 64 & 32 \\
Adam learning rate & 5e-6 & 1e-4 \\
Adam $\epsilon$ & 1e-5 & 1e-5  \\
PPO max gradient norm & 0.5  & 0.5  \\
PPO value clipping & yes  & yes \\
return normalisation & yes & no \\
value loss coefficient & 0.5 & 0.5 \\
student entropy coefficient &  & 0.0 \\
generator entropy coefficient &  & 0.0 \\

\addlinespace[10pt]
\emph{PLR} & & & \\
Scoring function &  & $\ell_1$ value loss \\
Replay rate, $p$ &  & 1.0 \\
Buffer size, $K$ &  & 512 \\
Prioritisation, &  & rank \\
Temperature, &  & 0.1 \\
Staleness coefficient, $\rho$ &  & 0.3 \\

\addlinespace[10pt]
\emph{RPLR} & & & \\
Scoring function, & & positive value loss \\
Replay rate, $p$ &  & 0.5 \\
Buffer size, $K$ &  & 4000 \\

\addlinespace[10pt]
\emph{ACCEL} & & \\
Edit rate, $q$ & & 1.0 \\
Replay rate, $p$ & & 0.8 \\
Buffer size, $K$ & & 4000 \\
Edit method, & & random  \\
Levels edited, & & easy \\

\addlinespace[10pt]
\emph{SSED} & & & \\
Replay rate, $p$ & & 1.0 \\
Scoring function support, & & dataset \\
Staleness support, &  & dataset \\
Secondary Scoring function, & & $\ell_1$ value loss \\
Secondary Scoring function support, & & buffer \\
Secondary Temperature, & & 1.0 \\
Mixing coefficient, $\eta$ & & linearly increased from 0 to 1 \\

\bottomrule 
\end{tabular}
}
\end{center}
\end{table}

\begin{table}[!htb]
\caption{Hyperparameters used for pre-training the VAE.}
\label{table:hyperparams_vae}
\begin{center}
\scalebox{0.88}{
\begin{tabular}{lccc}
\toprule
\textbf{Parameter} & \\
\midrule
\emph{VAE} & \\
$\beta$ & 0.0448 \\
layout head reconstruction coefficient & 0.04 \\
start and goal heads reconstruction coefficients & 0.013 \\
number of variational samples & 1 \\
Adam learning rate & 4e-4 \\
Latent space dimension & 1024 \\
number of encoder GCN layers & 4 \\
encoder GCN layer dimension & 12 \\
number of encoder MLP layers (including bottleneck layer) & 2 \\
encoder MLP layers dimension & 2048 \\
encoder bottleneck layer dimension & 256 \\
number of decoder layers & 3 \\
decoder layers dimension & 256 \\

\bottomrule 
\end{tabular}
}
\end{center}
\end{table}

\end{document}